\documentclass{article}
\usepackage{arxiv}
\usepackage{graphicx}
\usepackage[T1]{fontenc}
\usepackage[utf8]{inputenc}
\usepackage{amsfonts}
\usepackage{nicefrac}
\usepackage{hyperref}
\usepackage{array}
\usepackage{rotating}
\usepackage{float}
\usepackage{textcomp}
\usepackage{multirow}
\usepackage{amssymb}
\usepackage{amsmath}
\usepackage{stackrel}
\usepackage{subscript}
\usepackage[caption=false]{subfig}
\usepackage{booktabs}
\usepackage{comment}

\providecommand{\tabularnewline}{\\}
\floatstyle{ruled}
\newfloat{algorithm}{tbp}{loa}
\floatname{algorithm}{\protect\algorithmname}

\usepackage{algorithmic}
\floatname{algorithm}{Algorithm}
\usepackage{url}
\usepackage{float}

 \AtBeginDocument{%
 \abovedisplayskip=-10pt
 \abovedisplayshortskip=-10pt
 }

\graphicspath{{Images/}}

\begin{document}


\title{Revisiting Data Complexity Metrics Based on Morphology for Overlap and Imbalance: Snapshot, New Overlap Number of Balls Metrics and Singular Problems Prospect}


\author{Jos{\'e} Daniel Pascual-Triana\thanks{Andalusian Institute of Data Science and Computational Intelligence (DASCI), University of Granada, Granada, 18071, Spain.}\\
        jdpascualt@ugr.es \and
        David Charte \and
        Marta Andr{\'e}s Arroyo \and
        Alberto Fern{\'a}ndez \and
        Francisco Herrera\\
}





\date{Received: date / Accepted: date}
\maketitle

\begin{abstract}
Data Science and Machine Learning have become fundamental assets for companies and research institutions alike. As one of its fields, supervised classification allows for class prediction of new samples, learning from given training data. However, some properties can cause datasets to be problematic to classify.

In order to evaluate a dataset a priori, data complexity metrics have been used extensively. They provide information regarding different intrinsic characteristics of the data, which serve to evaluate classifier compatibility and a course of action that improves performance. However, most complexity metrics focus on just one characteristic of the data, which can be insufficient to properly evaluate the dataset towards the classifiers' performance. In fact, class overlap, a very detrimental feature for the classification process (especially when imbalance among class labels is also present) is hard to assess.

This research work focuses on revisiting complexity metrics based on data morphology. In accordance to their nature, the premise is that they provide both good estimates for class overlap, and great correlations with the classification performance. For that purpose, a novel family of metrics have been developed. Being based on ball coverage by classes, they are named after Overlap Number of Balls. Finally, some prospects for the adaptation of the former family of metrics to singular (more complex) problems are discussed.



\end{abstract}


\section{Introduction}\label{sec:introduction}


Automatic classification is an important area of Machine Learning \cite{Shwartz_2014}, given its usefulness in many contexts of everyday life \cite{ben-israel_impact_2020,Barboza2017405,martinez_torres_review_2019}. Supervised classification is the section of those problems where the sample instances have a prior labelling, which can be used to learn the pattern that leads to classifying new instances \cite{alpaydin_machine_2016}. To successfully address this task, it is advisable to acknowledge the properties of the problem. As such, experience acquired in similar scenarios may lead to ease the configuration of the global system, and thus to improve the predictive capabilities of the model via a meta-learning approach \cite{HKV2019}. 


Considering the former issue, datasets can present innate characteristics that help or hinder their performance. In order to obtain such information, many data complexity metrics have been developed over the years \cite{lorena_how_2019}, each of them centred on different characteristics: class overlap (whether different classes share an area \cite{orriols-puig_documentation_2010}), class imbalance (whether some classes present more data \cite{Fernandez_learning_2018}) or data neighbourhoods (what kind of boundaries the data has \cite{Lorena_analysis_2012}), among others. Moreover, these metrics can estimate the classification performance \cite{moran_2016} or flag the need for specific preprocessing on the data \cite{luengo_addressing_2011,Krawczyk2019601}, transforming and/or summarizing the problem before the classification \cite{Ahmed2019249}, in order to improve the results beforehand.

From those features, overlap and imbalance usually have great impact on classification performance \cite{alejo_hybrid_2013}. The former creates zones of dubious classification, whereas the latter can cause algorithms to ignore the classes with fewer elements. Even though their effects can be shallow when separate, when simultaneously present the consequences get compounded \cite{Fernandez_pareto_2017}. Specific techniques are already available to help tackle these situations \cite{Vuttipittayamongkol_2020,alshomrani_proposal_2015}, but their effect would still linger, so using unmodified situations is relevant when evaluating classifiers. 

Most of the current complexity metrics may fail to realise what is actually harming the classification, as they are designed to focus on a single property of the data. This issue suggests that metrics with broader scopes might perform better. Taking the former into account, the hypothesis of this paper is that data morphology, which observes a combination of class imbalance and overlap, as well as neighbourhood features, could prove beneficial. To that end, new morphology metrics will be proposed and their performance will be compared to that of the existent complexity metrics,  with particular interest on estimating overlap.

The novel complexity metrics are named ``Overlap Number of Balls'' (ONB) and check how hard to cover using class-dependent balls a dataset is. Since those balls can only include points from one class, covering areas with overlap takes more balls with smaller radii. Identifying how complex the boundaries between the classes are and combining both the local and global degrees of mixing, the ONB metrics should give good estimations of how complex a dataset is. 

Data complexity metrics have been widely studied in standard learning problems. However, to the best of our knowledge, very little contribution has been made regarding the use of data complexity metrics on the so-called ``singular problems'' \cite{charte_snapshot_2019}. These are known as those learning scenarios that might have more complex inputs (such as groups of instances, namely multiple-instance learning \cite{herrera_multiple_2016}) and outputs (such as multiple labels per instance, namely multi-label classification \cite{herrera_multilabel_2016}). In accordance with the former, most complexity metrics are not directly compatible when the structure of the problem changes. Despite that, their utilisation could prove just as useful as with the standard problems, especially when studying overlap. To cover that need, in this research work some guidelines will be provided for ONB's adaptation to singular problems.

Considering all the formerly discussed, the main contributions of this study are fourfold:

\begin{enumerate}
    \item To revisit morphology-based complexity metrics, extending what can be found in \cite{lorena_how_2019}.
    \item To provide new complexity metrics based on morphological features, namely the ONB metrics.
    \item To analyse how data morphology provides a good estimate of the classification results that can be obtained for a dataset.
    \item To give insight into how to apply morphology metrics on singular classification problems \cite{charte_snapshot_2019}. 
\end{enumerate}

For that purpose, the relationships between complexity metrics and the classification efficiency are checked. Aiming to avoid bias to specific learning schemes, different paradigms of classifiers have been selected, i.e. instance-based \cite{mazurowski_comparative_2011}, decision trees \cite{anuradha_self_2014}, and Bayesian models \cite{rodriguez_bayesian_2015}. Additionally, the performance is measured using two complementary metrics, namely the area under the ROC curve and the geometric mean. In the aforementioned experimental framework, both state-of the-art and newly proposed ONB metrics were used. In order to provide a controlled environment for the project, several artificial datasets with known theoretical overlap and imbalance values and different numbers of classes were created. The study is then completed with real benchmark problems, to acknowledge the applicability to real situations. Supplementary information regarding this paper, including the full results and the explanations of all the used complexity metrics, can be found in this web repository \footnote{\url{https://github.com/jdpastri/morphology-metrics}}.

In order to accomplish the former goals, the rest of this paper is organised with the following structure. Section \ref{sec:fundamentals} introduces the general information regarding classification problems and their base typology. Section \ref{sec:describing} introduces data overlap and performs a revision of the state-of-the-art complexity metrics. Section \ref{sec:ONB} presents the novel ONB morphology-based overlap metrics. Section \ref{sec:framework} describes the experimental analysis of this study. Section \ref{sec:results} discusses the experimental results. Section \ref{sec:adap} adds suggestions for the adaptations of the ONB algorithms to a most types of singular problems. Finally, Section \ref{sec:remarks} presents the concluding remarks. For the sake of completeness, the Appendix describes the non-morphology-based state-of-the-art complexity metrics (also available online\footnotemark[1]). 

\section{Preliminaries on Classification Problems}\label{sec:fundamentals}

Due to classification problems being multidisciplinary, there is a wide range of different paradigms regarding both the input and the expected output \cite{aggarwal_data_2014}, \cite{charte_snapshot_2019}. Therefore, it would be advisable to present some of the most notable situations that can be found when addressing real-world problems. First, the baseline classification task is described in Section \ref{subsec:preliminaries}. Then, the focus changes to modifications of the general paradigm, such as multiclass learning (Section \ref{subsec:multiclass}) and classification with imbalanced data (Section \ref{subsec:imbalanced}).

\subsection{Binary classification problems}\label{subsec:preliminaries}

The baseline classification problem is binary, that is, it involves data with only two possible classes, and where both labels are almost equally represented. The aim is, as expected, to be able to discern the class of some new data elements given a (normally bigger) training set. To carry out the modelisation of the problem, one or more classification algorithms are chosen and executed over the training data. These models are used for the assignation of labels of future data. A binary classification model, therefore, can be represented as $f:X\rightarrow Y$, a function which takes input data from a set of possible feature vectors $X\subset \mathbb{R}^n$ and outputs a prediction from the two-element set $Y=\{0,1\}$. The classification problem consists in learning this model from a finite set of examples $S=\{(x,y)\in X\times Y\}$.

The goodness of the model is then evaluated following a chosen metric, usually the accuracy. This metric is the rate of the correctly classified data instances divided by the total number of instances. High accuracy values indicate a good performance of the algorithm when modelling the problem.

To improve the classification results, aside from changing the learning algorithm, there are two main courses of action:

\begin{enumerate}

\item \emph{Fine-tuning the hyper-parameters of the algorithm}. This consists on adapting the parameters of the algorithm to the problem at hand, so that the classification can be more lenient or more severe in certain areas of the data space \cite{luo_2016}. For the parameter choices there are multiple options, such as using grid search \cite{feurer_2019} or random search \cite{bergstra_2012}. However, caution should be taken, as too much fine-tuning can lead to overfitting \cite{hutter_2019}.
\item \emph{Smart Data preprocessing}. Base data and its distinctive features can greatly influence the classification results. If the data presents features that will harm the classification process, executing some preprocessing can lead to more informative and meaningful data (currently known as Smart Data \cite{luengo_2020}) and thus, improve the performance of the classifier over the test set \cite{garcia_tutorial_2016}. However, some caution is advised, as modifying the dataset too much might create models that do not resemble reality.

\end{enumerate}

\subsection{Multiclass learning}\label{subsec:multiclass}

This kind of classification problem has more than two possible output classes for the points in the studied dataset \cite{gupta2014training}.  More formally, a multiclass problem consists in finding a classifier $f:X\rightarrow Y$ where $Y$ is finite, but holds more than two elements. This implies the need to obtain the regions spanning each of the classes, either by pairwise comparisons or all simultaneously, and thus creates more complex boundaries \cite{GALAR2014135}.

Furthermore, some algorithms are not suitable for the analysis of more than two classes at the same time. This means that, sometimes, these problems need to be transformed into multiple binary problems. Two main schemes are possible so that the algorithm can appropriately handle them, either for each pair of classes (one vs. one), or for each class against the rest (one vs. all) \cite{GalarFTSH11}. Another option to face this complication is to adapt the algorithms so that they are able to study multiple classes at a time.

Lastly, it should be noted that, for multiclass learning, evaluating the classification performance via a confusion matrix \cite{aggarwal_data_2014} is especially important. It separates the hit and fail rates by class pairs, which provides more insight on which classes are harder to classify than a general approach and, showing which classes get mixed, allows for the choice of better (targeted) preprocessing.

\subsection{Imbalanced classification}\label{subsec:imbalanced}

For many classification problems, the class distribution is imbalanced \cite{Fernandez_learning_2018}. The structure of the classification is very similar to the general case, except for the fact that one of the classes, called positive class, has a substantially lower amount of data points \cite{Lopez_insight_2013}. The other class receives the denomination of negative class. Expressed mathematically, this means that for dataset $S\subset X\times Y$ there exist $y_-, y_+\in Y$ verifying $\lvert\{(x,y_+)\in S\}\rvert << \lvert\{(x,y_-)\in S\}\rvert$.


Thus, if the same general procedure were followed, the class with the most data would receive preference by the algorithms in areas of the data space where the assignation was uncertain, worsening the classification of the other classes \cite{leevy_survey_2018}. In cases of high imbalance, it is also possible for the positive class to be outright ignored, which is a regrettable outcome.

For this reason, the metrics normally utilised to evaluate the performance of classification algorithms, such as the classification accuracy, recall, or specificity on their own, are less representative  \cite{aggarwal_data_2014}. Because of this, other measures need to be used that take into account the classification success of both the positive and negative classes simultaneously, but not as a whole . This includes metrics such as F1, the area under the ROC curve (AUC), the geometric mean of the true positive and negative rates, or other combinations of the former \cite{Luque2019216}.

In order to avoid the classification problems that would arise, the data of imbalanced datasets is usually preprocessed to balance the classes \cite{prati_2015}.  For this purpose, there are multiple courses of action. On the algorithm level, its parameters can be tuned to penalise the misclassification of the positive class; on the data level, balancing strategies are based on either the addition of new positive cases \cite{bellinger_2019}, the elimination of negative cases \cite{Mahani_2019} or the combination of both strategies \cite{serrano_2018}, improving the boundaries or eliminating noisy instances.

Alternately to dataset transformations, cost-sensitive learning applies different costs to each class, so that learners will pay special attention to positive instances, thus compensating the scarcity of samples of the minority class \cite{Li_cost_2017}. 

\section{Snapshot on Morphology-based Complexity Metrics}\label{sec:describing}

This research paper focuses on the use of data complexity metrics and how they can allow for the description of a dataset and the estimation of the degree of overlap among its classes. For that purpose, it is necessary to present an overview that explains what they are and how they have been used historically.

This section starts with an introduction to data complexity metrics, stating their purpose and the main works regarding their current situation (Section \ref{subsec:intro_cm}). Then, the state-of-the-art metrics are indicated in Section \ref{subsec:state}. Finally, those related to data morphology (the neighbourhood metrics) are explained in Section \ref{subsec:neighbourhood}. 

\subsection{Introduction to data complexity metrics and overlap}\label{subsec:intro_cm}

Every dataset has its own intrinsic characteristics that differentiate it from others and that facilitate or hinder its study and classification \cite{wojciechowski_difficulty_2017}, \cite{Das2018HandlingDI}. Data complexity metrics serve to evaluate specific features, such as the overlap among classes, the structure of the neighbourhood of its examples, the linear separateness, the dimensionality of the data, its structure, the class balance, network properties or morphology, among others.

Many researchers take the complexity metrics presented by Tim Kam Ho and Mitra Basu in \cite{ho_complexity_2002} as a basis. Their set of metrics already studied overlap, separability, geometry, topology and density of manifolds. Further complexity metrics have been compiled in \cite{lorena_how_2019}, which reviews the state-of-the-art complexity metrics. It follows mostly a theoretical approach for their description and presents a ready-to-use implementation of them, but does not explore which ones perform better and under which circumstances.

To better study the relevance of complexity metrics, Scopus \cite{scopus_document_nodate} (Elsevier's bibliography database) was queried for the computation or engineering articles whose titles included ``metrics'' and ``data complexity'', by using the following query:

{
\begin{verbatim}
TITLE-ABS-KEY ("metrics") 
AND TITLE-ABS-KEY ("data complexity") 
AND (LIMIT-TO (DOCTYPE, "ar"))   
AND (LIMIT-TO (SUBJAREA, "COMP")
  OR LIMIT-TO (SUBJAREA, "ENGI")).
\end{verbatim}}

\begin{figure}[!ht]
\begin{centering}
\includegraphics[width=\linewidth]{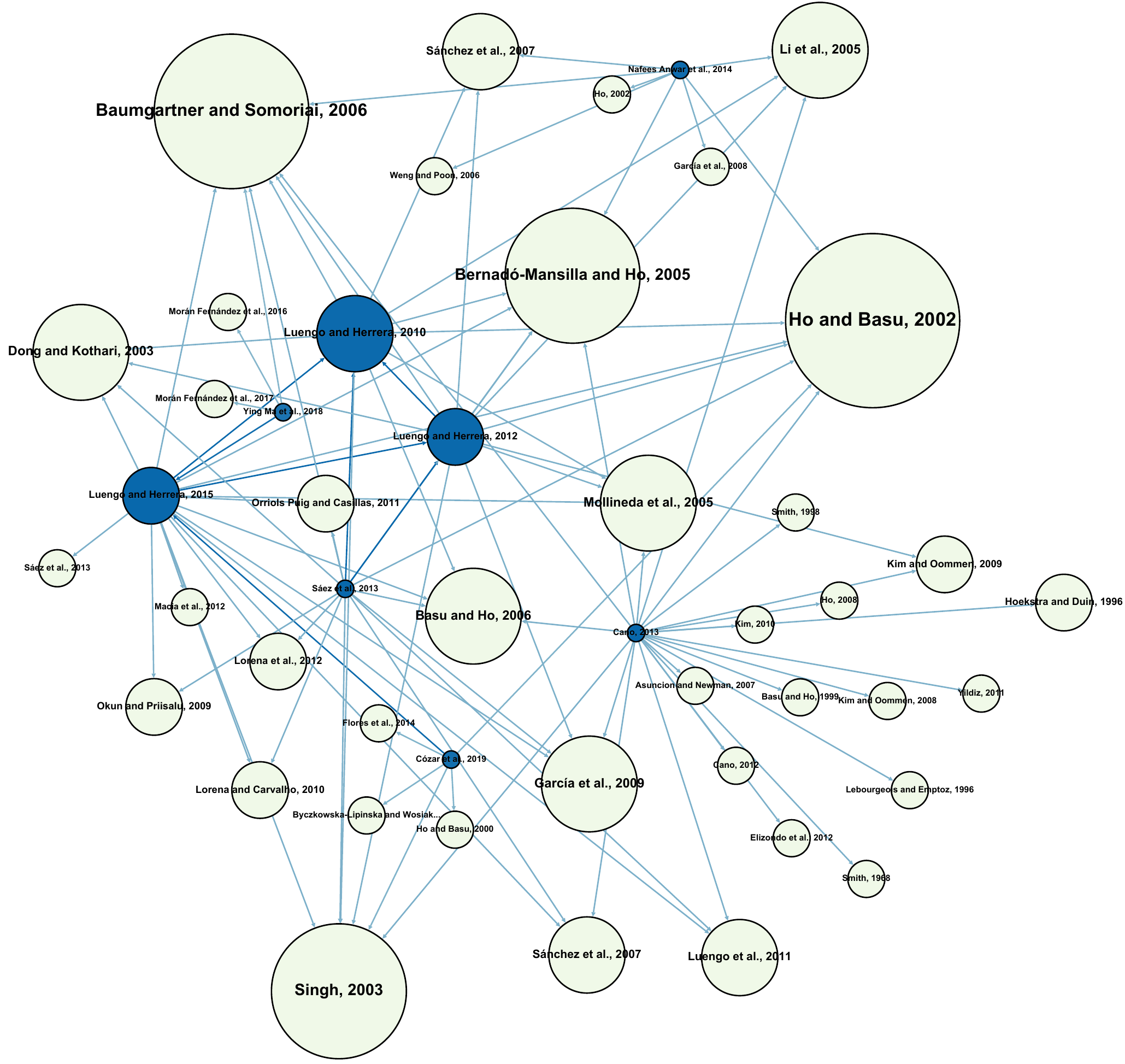}
\end{centering}
\caption{\label{fig:red}Network of articles on data complexity metrics, created from the Scopus query.}
\end{figure}

From the 25 articles (as of May 2020), the 8 examples that actually used or introduced data complexity metrics were selected and, using their references from this field, the article network in Figure \ref{fig:red} was created. The blue nodes form the starting set of 8 papers and the directed edges connect them to the papers they cite (either new, grey nodes or other blue nodes). The size of the nodes corresponds to their in-degree (how many times they have been cited).

Thus, from that selection, most papers cite Ho and Basu, 2002 \cite{ho_complexity_2002}, Baumgartner and Somorjai, 2006 \cite{baumgartner_data_2006}, Bernadó-Mansilla and Ho, 2005
\cite{bernado-mansilla_domain_2005} and Singh, 2003 \cite{singh_multiresolution_2003}. These papers have therefore had an impact on the development and use of complexity metrics as a whole, presenting many of the base metrics from which others have been derived.

Most of the measures in those papers allow for an estimation of how intertwined the different classes are. Most often, the more overlapped the classes of a dataset are, the worse the classification results. Therefore, studying the complexity of a dataset before applying algorithms for its classification can provide a reference for the quality of the results.

This has been the focus of multiple of those works, such as \cite{ma_data_2018}, where several complexity metrics are used on software defect detection datasets to obtain a correlation wih the results of KNN, C4.5 and Naïve Bayes, or in \cite{cano_analysis_2013}, where the relationship between some complexity metrics and the classification results from C4.5, 3NN and SVM is checked over several datasets.

What is more, data complexity metrics can even be used to determine which classification algorithms would be more suitable for that particular dataset. In \cite{luengo_domains_2010} fuzzy rules from the ranges of a set of complexity metrics that infer good or bad behaviours when using SVM and ANN are obtained. Other similar works have been developed by the same authors with respect to other measures and algorithms, as can be seen in \cite{luengo_shared_2012} or in \cite{luengo_automatic_2015}, where the value intervals of the metrics are studied with respect to KNN, SVM and C4.5. Other works include \cite{cozar_metahierarchical_2019}, where a fuzzy decision rule system is developed in order to decide which classifier to use in a meta-classifier, learnt from the values of several complexity metrics of a diverse collection of datasets.

Moreover, complexity metrics can also signal the need for preprocessing steps on the data, which could eventually improve the performance of the algorithm, such as in \cite{saez_predicting_2013}, where complexity metrics are used to develop decision rules that determine whether noise filtering is necessary before classifying a dataset.

\subsection{State-of-the-art complexity metrics}\label{subsec:state}

This paper aims to revisit the complexity metrics used to detect class overlap and how they relate to the classification results. In this sense, the methods can be categorised depending on the characteristics of the data they focus on:

\begin{itemize}
    \item \emph{Feature overlap metrics} gauge how features can discern the classes.
    \item \emph{Linearity metrics} study the separability of classes by hyperplanes.
    \item \emph{Neighbourhood metrics} check the boundaries among classes from each instance's surroundings.
    \item \emph{Dimensionality metrics} inspect data sparsity.
    \item \emph{Class Balance metrics} care about whether classes are equally represented
    \item \emph{Network Properties metrics} focus on the graph properties of the data.
\end{itemize}

In particular, the most prominent complexity metrics, which are included in \cite{lorena_how_2019}, were selected as the state-of-the-art. Table \ref{tab:Recordatorio-de-las} summarises those metrics.

\begin{table}[!ht]
\caption{State-of-the-art metrics from \cite{lorena_how_2019}.
\label{tab:Recordatorio-de-las}}

\centering{}
\begin{tabular}{ccc}
\toprule
Type of metric & \multicolumn{2}{c}{Metric name}\tabularnewline
\midrule
 
\multirow{3}{*}{Feature Overlap} & F1 & F1v\tabularnewline
 
 & F2 & F3\tabularnewline

 & F4 & \tabularnewline
\midrule
\multirow{2}{*}{Linearity} & L1 & L2\tabularnewline

 & L3 & \tabularnewline
\midrule
\multirow{3}{*}{Neighbourhood} & N1 & N2\tabularnewline

 & N3 & N4\tabularnewline

 & T1 & LSC\tabularnewline
\midrule
\multirow{2}{*}{Dimensionality} & T2 & T3\tabularnewline

 & T4 & \tabularnewline
\midrule
\multirow{1}{*}{Class Balance} & C1 & C2\tabularnewline
\midrule 
\multirow{2}{*}{Network Properties} & Density & ClsCoef\tabularnewline

 & Hubs & \tabularnewline
\bottomrule
\end{tabular}
\end{table}

From those categories, neighbourhood metrics are clearly related to data morphology, the focus of this paper. For the sake of completeness, a summary description for the remainder
studied metrics can be found in the Appendix of this manuscript. Should it be required, further information about these metrics can be found in \cite{lorena_how_2019}.



\subsection{Neighbourhood metrics}\label{subsec:neighbourhood}

These metrics aim to discern the boundaries between the classes or their structure, studying the neighbourhoods of the points. These six metrics are the most prevalent:
\begin{itemize}
\item N1: it is the Fraction of Borderline Points of the Minimum Spanning Tree, a tree graph generated using the instances (vertexes) and the distances between them to create and weigh the edges. The borderline points are those that have at least one edge that connects them to a point of a different class. N1 is the ratio of these points and the total number of points. For the graph $MST(V,E)$ and where $y_{i}$ is the class of instance $x_{i},\,i=1,...,n$, N1 is given by Equation \ref{n1}. Greater values indicate greater complexity. 

\begin{center}
\begin{equation}
N_{1}=\frac{1}{\left\lvert V\right\rvert}\left\lvert\left\{x_{i}\in V\,:\,\exists(x_{i},x_{j})\in E,\:y_{i}\text{\ensuremath{\neq}}y_{j}\right\}\right\rvert\label{n1}
\end{equation}
\end{center}

\item N2: this is the Ratio of Intra/Extra-Class Nearest Neighbour Distance, which compares the distances inside a class with those between classes. For this, the division of the sum of the 1NN distances among the points from one class and the sum of the 1NN distances of points from that class to points of a different class is computed. N2 is a transformation of that division which, 
for a dataset $S$ with $n$ instances, $\operatorname{1NN^i}(x)$ being the intra-class nearest neighbour to $x$ and $\operatorname{1NN^e}(x)$ the extra-class nearest neighbour, is given by Equation \ref{n2}.
The bigger that ratio (and therefore N2), the closer the classes, which often signals greater complexity.

\begin{center}
\begin{equation}
N_2 = \frac{r}{1+r}\mbox{, where }
r=\frac{\stackrel[(x,y)\in S]{}{\sum} d(x, \operatorname{1NN^i}(x))}{\stackrel[(x,y)\in S]{}{\sum} d(x, \operatorname{1NN^e}(x))}
\label{n2}
\end{equation}
\end{center}

\item N3: this is the Error Rate of the Nearest Neighbour Classifier, or the ratio of the misclassified points using a leave-one-out 1NN over all instances. For a dataset $S$ with $n$ samples, where $\operatorname{1NN'}(x)$ is the leave-one-out nearest neighbour of sample $x$ and the class of a sample is retrieved via function $c$, N3 can be written as in Equation \ref{n3}. The higher the value, the more complex the boundaries and the dataset will be.

\begin{center}
\begin{equation}
N_{3}=\frac{1}{n}\left\lvert\left\{(x,y)\in S:\,c(\operatorname{1NN'}(x))\neq y\right\}\right\rvert\label{n3}
\end{equation}
\end{center}

\item N4: it is the Non-Linearity of the Nearest Neighbour Classifier. For this metric, a set  of new points $S'\subset X\times Y$ is generated by interpolation of pairs of points that share a class. These new points are then classified using nearest neighbours, where the initial data is the training set. N4 is the ratio of the misclassified interpolated points. Given dataset $S'$ of interpolated samples, where $\operatorname{1NN_{orig}}(x)$ is the nearest neighbour from the initial set of samples $S$, N4 can be written as in Equation \ref{n4}. Higher values indicate a more complex dataset.

\begin{center}
\begin{equation}
N_{4}=\frac{1}{\left\lvert S'\right\rvert}\left\lvert\left\{(x,y)\in S'\,:\,c(\operatorname{1NN_{orig}}(x))\neq y\right\}\right\rvert\label{n4}
\end{equation}
\end{center}

\item T1: this is the Fraction of Hyperspheres Covering Data, from \cite{ho_complexity_2002}. For this metric, hyperspheres are generated, centred on each data point. Their radii are the distances from those instances to their closest of a different class. Then, the hyperspheres that are completely contained within others are eliminated. T1 is the ratio between the number of the chosen hyperspheres and the total points which, for a dataset with $n$ instances, is shown in Equation \ref{t1}. Higher values imply a harder coverage and a more complex dataset.

\begin{center}
\begin{equation}
T_{1}=\frac{\mathit{Hyperspheres}}{n}\label{t1}
\end{equation}
\end{center}

\item LSC: this is the Local Set Average Cardinality, from \cite{leyva_set_2015}. The local set of an instance consists of the data points from its class that are closer than those of other classes. Its cardinality can indicate the closeness of that instance to the class boundaries, since points near them would have close neighbours from a different class, and their local set would thus hold few neighbours from their same class. LSC is derived from the ratio between the mean of those cardinalities and the total instances, which for a dataset $S$ with $n$ instances where the nearest neighbour from a different class can be found by function $\operatorname{1NN^e}$, is shown in Equation \ref{lsc}. Higher LSC indicates less complexity.

\begin{center}
\begin{equation}
LSC=1-\frac{1}{n^{2}}\stackrel[(x,y)\in S]{}{\sum}\left\lvert\left\{(x',y')\in S\,:\,d(x,x')<d(x,\operatorname{1NN^e}(x))\right\}\right\rvert\label{lsc}
\end{equation}
\end{center}
\end{itemize}

\section{Overlap Number of Balls (ONB): Complexity Metrics Based on Morphological Features}\label{sec:ONB}

As indicated in the introduction of the paper, metrics that focus too much on a single characteristic can fail to explain the intricacies of a dataset, and thus their relationship with the classification results can be limited. In order to extend the research to other complexity metrics with a broader point of view, which should provide a better data characterisation, some original metrics based on the morphology of the data will be also proposed in this study: the Overlap Number of Balls (ONB) family. 

This section starts with the explanation of the P-CCCD algorithm, which is used for the creation of a ball coverage set for ONB (Section \ref{subsec:PCCCD}). Then, ONB complexity metrics will be presented (Section \ref{subsec:ONB}).

\subsection{Preliminaries: the P-CCCD algorithm}\label{subsec:PCCCD}

The Pure Class Cover Catch Digraph (P-CCCD) is a greedy algorithm from \cite{manukyan_classification_2016} for the creation of a ball coverage set. It chooses, for every step, the hypersphere centred on a point of the studied class that includes the most points of that same class without including points of another class, until all points of the target class are covered. This will not always provide the optimal covering (since doing so would be a NP-hard task), but it is a good approximation. Its pseudocode is available is presented in Algorithm \ref{alg1}, where $d_{i,j}$ is the distance between point $i$ and point $j$, $c(i)$ is the class of element $i$ and $c_{t}$ is the target class.

\begin{algorithm}[!ht]
\begin{algorithmic}[1]

\STATE  \textbf{input $d_{i,j}:i,j=1...n$}

\STATE  $\;\;\;\;\;U=\{i/c(i)=c_{t}\}$

\STATE  $\;\;\;\;\;V=\{i/c(i)\neq c_{t}\}$

\STATE  b=0

\STATE  \textbf{while} $U\neq\phi$

\STATE  \textbf{$\;\;$}G=\{\}

\STATE  \textbf{$\;\;$for} i in 1:n

\STATE $\;\;\;\;d=min_{j\in V}(d_{i,j})$

\STATE $\;\;\;\;P=\{u\in U/d_{i,u}<d\}$

\STATE $\;\;\;\;$\textbf{if} $|P|>|G|$

\STATE $\;\;\;\;\;\;$$G=P$

\STATE \textbf{$\;\;$end for}

\STATE  \textbf{$\;\;$}b=b+1

\STATE  \textbf{$\;\;U=\{u\in U$$\backslash$$u\notin G\}$}

\STATE  \textbf{end while}

\STATE  \textbf{return }b

\STATE  \textbf{end}

\end{algorithmic}

\caption{Pseudocode outline of P-CCCD with $\tau=1$.\label{alg1}}
\end{algorithm}

\begin{figure}[ht]
    \centering

    \subfloat[]{\label{ejbolasa}
    \includegraphics[width=.45\linewidth]{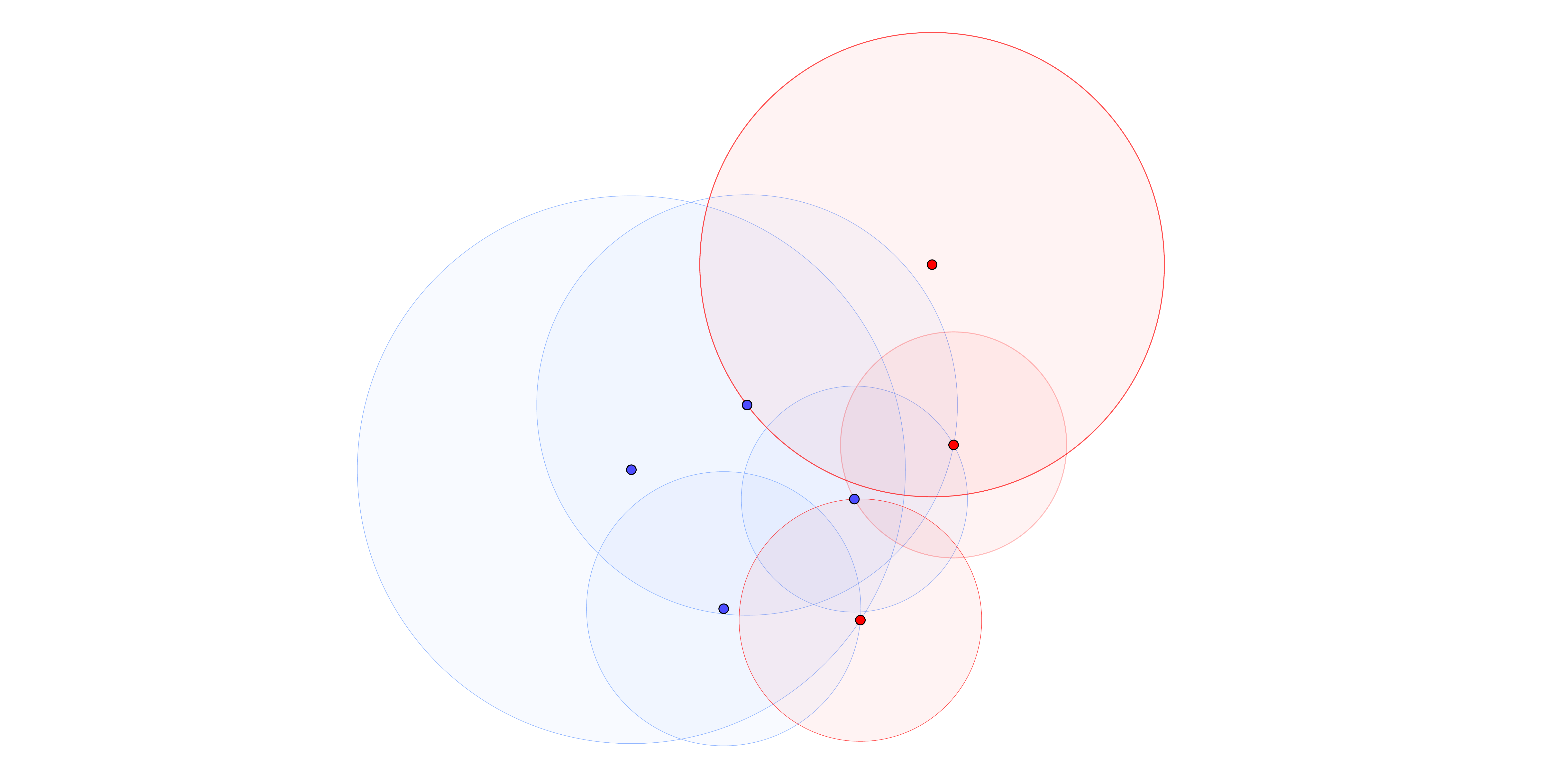}}
    \subfloat[]{\label{ejbolasb}
    \includegraphics[width=.45\linewidth]{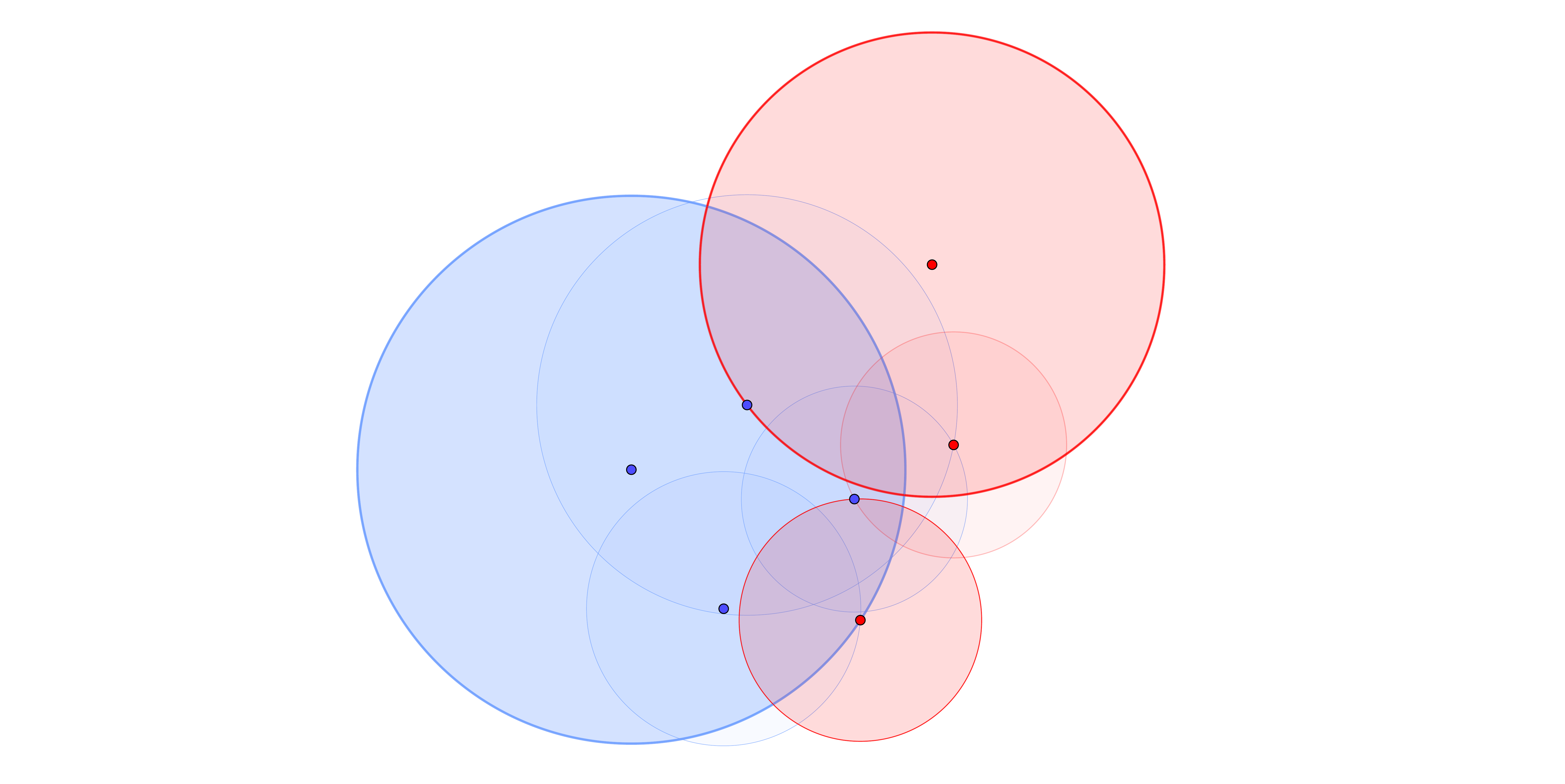}}

\caption{\label{ejbolas}Example of the P-CCCD ball coverage algorithm. First, the maximum radii for all instances are calculated (a), then the balls covering the most points of each class are iteratively chosen (b).}
\end{figure}

To better illustrate this pseudocode, Figure \ref{ejbolas} is presented. For every point, the biggest ball that does not include points of a different class is created, and only the one that includes the most same-class points is selected for each step, until all points are covered.

\subsection{ONB family of complexity metrics}\label{subsec:ONB}

The ONB complexity metrics are based in the class-dependent ball coverage of the data. Thus, they are affected by data overlap and the morphology of the class boundaries. However, since this coverage involves distance calculations, the distance metric also has a definite effect on the result. 

\begin{figure}[!ht]
\begin{centering}
\resizebox{0.4\textwidth}{!}{\includegraphics{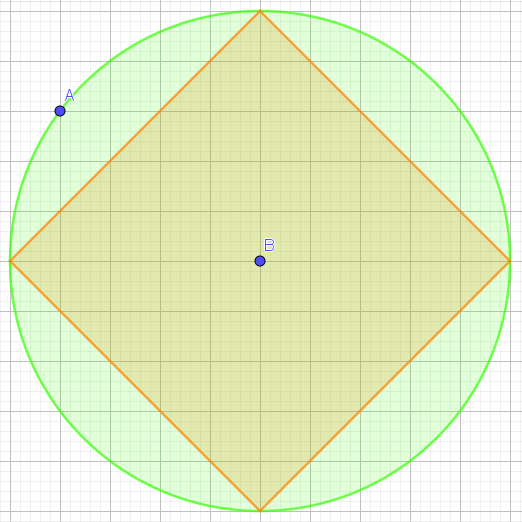}}
\par\end{centering}
\caption{\label{boleucman}Euclidean (green) and Manhattan (orange) balls with the same radius.}
\end{figure}

To avoid bias, two well-known distance metrics were chosen for ONB: the Euclidean distance and the Manhattan distance. The balls generated using them follow different structures and, as can be seen in Figure \ref{boleucman}, the Manhattan distance between two points is usually greater. This allows the analysis of different behaviours for overlap detection.

After obtaining the number of balls with the chosen distance metric for the coverage of each class, 2 types of metrics are proposed, depending on whether the class balance was considered: ONB\textsubscript{tot} and ONB\textsubscript{avg}.

\begin{itemize}

\item $ONB_{tot}$ does not differentiate by class. It is simply the ratio between the number of balls necessary to cover the points of all classes and the number of points of the dataset. Equation \ref{onb_tot} presents the metric, where $b_{i}$ is the number of balls for class $i$, $n_{i}$ is the number of elements of said class and $k$ is the number of classes.

\begin{center}
\begin{equation}
ONB_{tot}=\frac{\stackrel[i=1]{k}{\sum}b_{i}}{\stackrel[i=1]{k}{\sum}n_{i}}.\label{onb_tot}
\end{equation}
\par\end{center}

\item $ONB_{avg}$ differentiates by class. The ratio between the number of balls necessary to cover the points of a class and the number of points of that class is calculated and then averaged over the classes. Consequently, it considers whether some classes are more difficult to cover than others, which would raise the mean. Equation \ref{onb_avg} presents the metric, where $b_{i}$ is the number of balls for class $i$, $n_{i}$ is the number of elements of said class and $k$ is the number of classes.

\begin{center}
\begin{equation}
ONB_{avg}=\frac{\stackrel[i=1]{k}{\sum}\frac{b_{i}}{n_{i}}}{k}.\label{onb_avg}
\end{equation}
\par\end{center}

\end{itemize}

This $ONB_{tot}$ might be reminiscent of T1 (Section \ref{subsec:neighbourhood}), since both use ball/sphere coverage. However, due to its ball-coverage algorithm, $ONB_{tot}$ achieves a more minimal coverage and simpler boundaries between the classes. For instance, in Figure \ref{ejbolas}, 1 ball per point would have been necessary for T1 ($T1=1$), since no ball is completely included in another, and that is not a good representation of the complexity of the former problem.

Given the use of two different distance metrics and two distinct methods for result aggregation, the four proposed complexity metrics have been summarised in Table \ref{t7}.

\begin{table}[h]
\caption{\label{t7}New proposed metrics based on covering balls.}

\centering{}%
\begin{tabular}{ccc}
\toprule
{Distance} & {Type} & {Nomenclature}\tabularnewline
\toprule
\multirow{2}{*}{{Euclidean}} & {Total ($ONB_{tot}$)} & {$ONB_{tot}^{euc}$}\tabularnewline

 & {Average ($ONB_{avg}$)} & {$ONB_{avg}^{euc}$}\tabularnewline
\midrule 
\multirow{2}{*}{{Manhattan}} & {Total ($ONB_{tot}$)} & {$ONB_{tot}^{man}$}\tabularnewline

 & {Average ($ONB_{avg}$)} & {$ONB_{avg}^{man}$}\tabularnewline
\bottomrule 
\end{tabular}
\end{table}


\section{Experimental Framework}\label{sec:framework}

One of this study's aims is to evaluate the performance of the state-of-the-art complexity metrics, with emphasis on morphology metrics, both in terms of overlap estimation and relationship with the classification performance. For that purpose, a solid base of datasets is essential, which led to a twofold approach: creating artificial datasets with known theoretical overlap and imbalance to provide a controlled base environment, and the use of real datasets to contrast the results. Using different classifiers and measures for their performance over the datasets was also necessary to ensure the generality of the results.

This section is organised as follows. First, the description of the datasets that will be used in the study is shown in Section \ref{subsec:datasets}, detailing the creation of the artificial datasets and presenting the real datasets, chosen to corroborate the real applications of the metrics. This is followed by the classification methodology in Section \ref{subsec:methodology}, which presents the classifiers and their efficiency metrics. 

\subsection{Dataset description}\label{subsec:datasets}

In order to establish a controlled basis for the experimental framework, several artificial datasets were created, using as a basis simple plane figures: circles, rectangles and triangles. These shapes were chosen due to their intrinsic characteristics:
\begin{itemize}
\item Circles have non-linear boundaries, which are harder to grasp for some classifiers but, when two circles of the same size overlap, they have symmetry axes that can be perpendicular to the features.
\item Rectangles have linear boundaries, both in absence and presence of overlapping areas, so linear classifiers should be favoured when studying these datasets. Probabilistic classifiers could have a harder time modelling the problems.
\item Triangles have linear boundaries, but some of them are oblique, which is hard to grasp for some classifiers whose bases are perpendicular divisions. 
\end{itemize}
Using these shapes as a basis, 72 balanced datasets with different values of overlap were created. Their characteristics are compiled in Table \ref{t4-1}. 

\begin{table}[!ht]
\caption{\label{t4-1}Outline of the balanced artificial datasets.}

\centering%

\begin{tabular}{ccc}
\toprule 
{N.º of classes} & {Figures} & {Overlap (\%)}\tabularnewline
\toprule 
\multirow{3}{*}{{2 (binary)}} & {Circles} & \multirow{8}{*}{{0, 5, 10, 15, 20, 25, 30, 50, 100}}\tabularnewline

 & {Rectangles} & \tabularnewline

 & {Triangles} & \tabularnewline
\cline{1-2} 
\multirow{3}{*}{{3}} & {Circles by pairs} & \tabularnewline
 
 & {Circles together} & \tabularnewline
 
 & {Circles inside} & \tabularnewline
\cline{1-2} 
\multirow{2}{*}{{4}} & {Rectangles} & \tabularnewline

 & {Triangles} & \tabularnewline
\hline 
\end{tabular}

\end{table}

The overlap was calculated theoretically, from the areas encompassed by each class, in order to have direct control of the characteristics of the datasets. Whenever possible,
these overlap percentages are kept the same for every pair of classes; that is, in every case but the ``circles inside'' datasets, where two classes are contained inside a third, and the 4-class triangle datasets, where adjacent and opposed classes have different degrees of overlap. 

For the datasets with 4 classes, additionally, the overlap percentage is recalculated using the binary percentage.

\begin{figure*}[h]
\centering

\includegraphics[width=.2\linewidth]{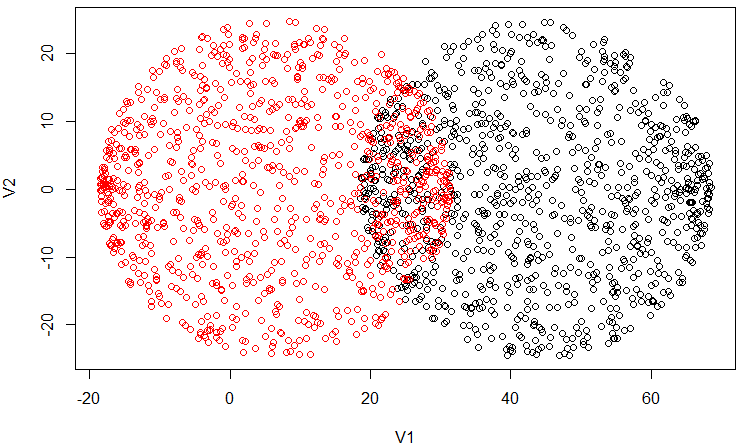}
\includegraphics[width=.2\linewidth]{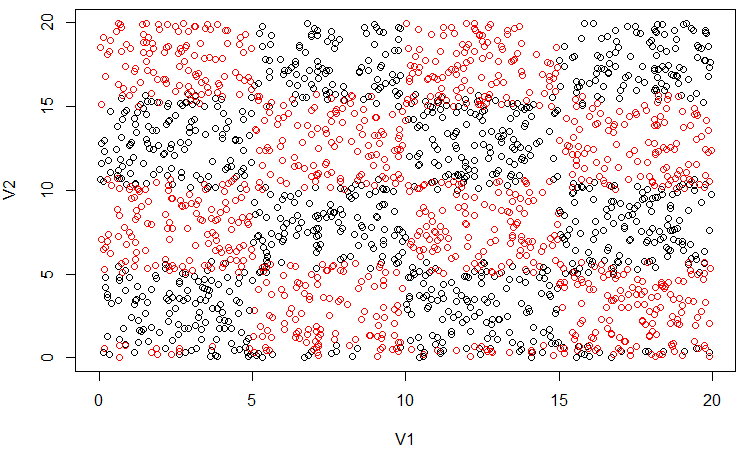}
\includegraphics[width=.2\linewidth]{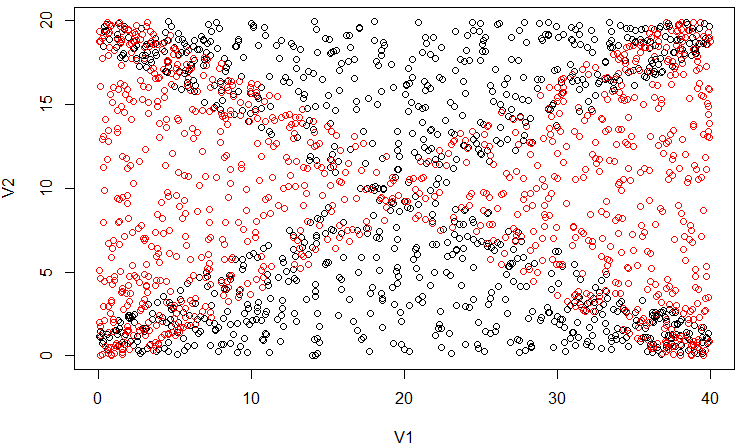}
\includegraphics[width=.2\linewidth]{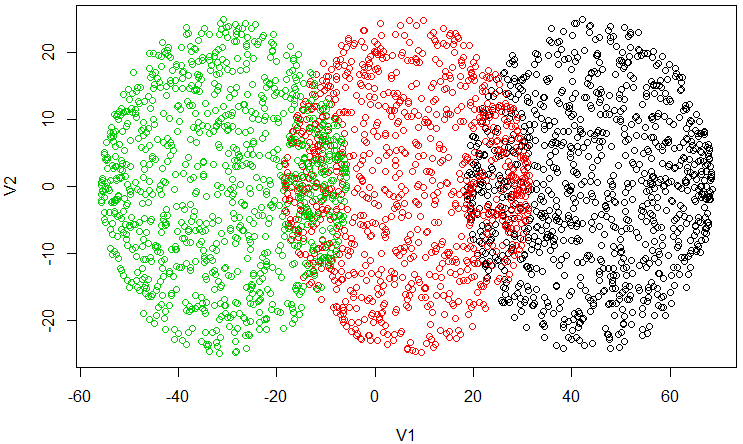}

\includegraphics[width=.2\linewidth]{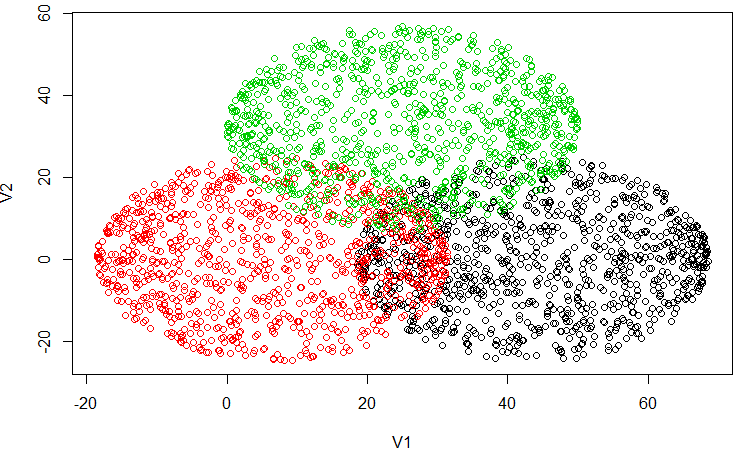}
\includegraphics[width=.2\linewidth]{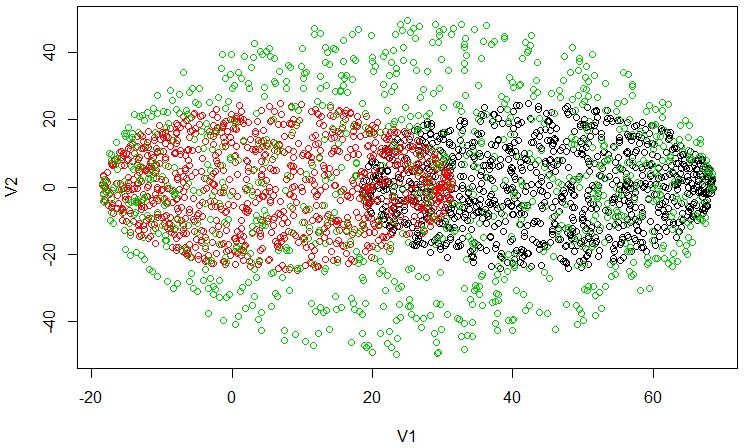}
\includegraphics[width=.2\linewidth]{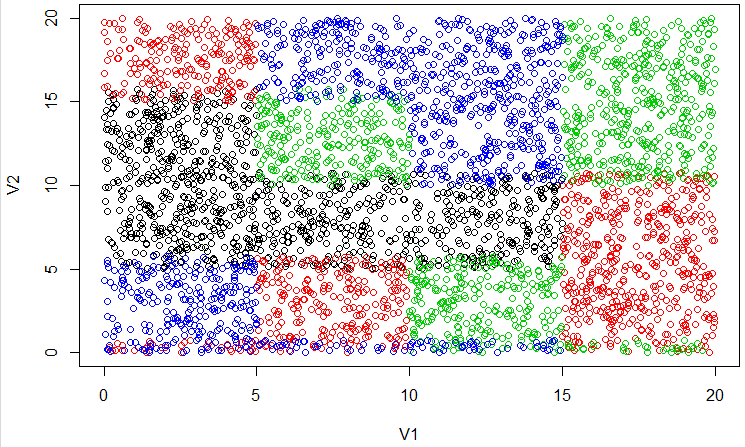}
\includegraphics[width=.2\linewidth]{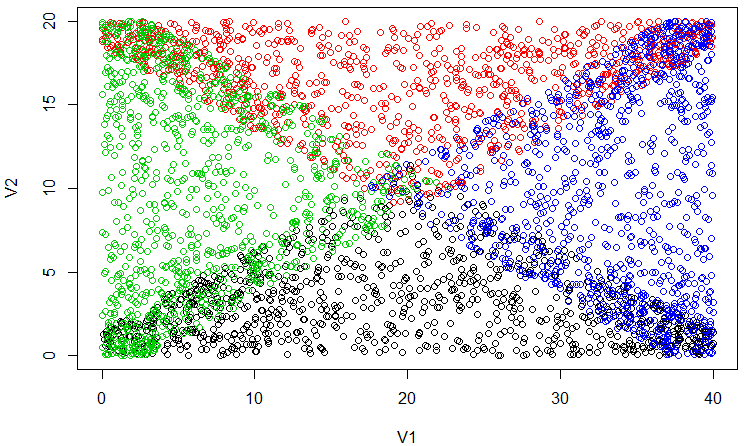}
\caption{Examples of the artificial datasets and how they overlap.\label{fig:ej_art}}

\end{figure*}

Some samples of those datasets have been chosen to illustrate their obtaining (see Figure \ref{fig:ej_art}). It must be acknowledged that the scale may distort the shapes of the circles.

\begin{figure}[h]
\begin{centering}
\includegraphics[width=.45\linewidth]{circulostripletes3cl15sol}%
\includegraphics[width=.45\linewidth]{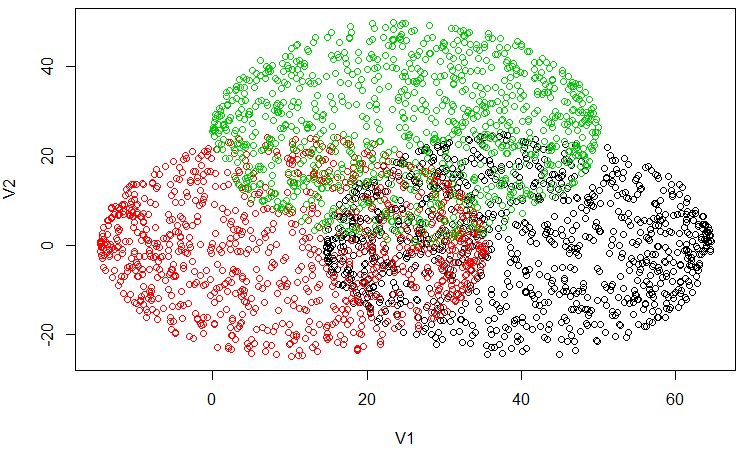}
\par\end{centering}
\caption{\label{dif_ov}Examples of a synthetic dataset with different overlap degrees.}

\end{figure}

Different degrees of overlap with the same dataset structure are shown
in Figure \ref{dif_ov}.

\begin{figure}[!h]
\begin{centering}
\includegraphics[width=.45\linewidth]{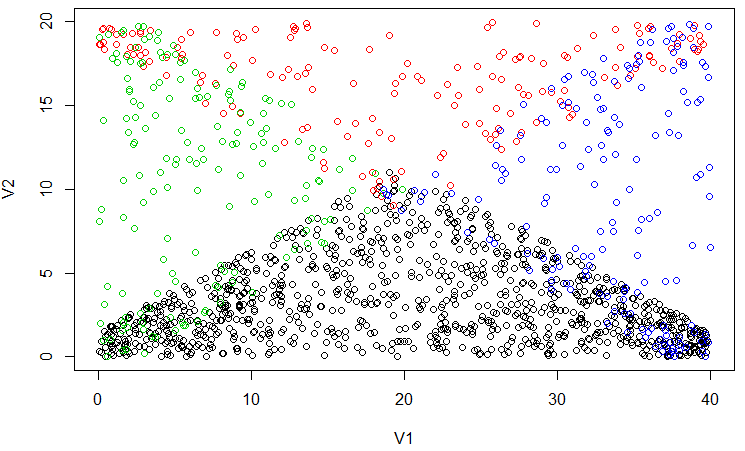}
\includegraphics[width=.45\linewidth]{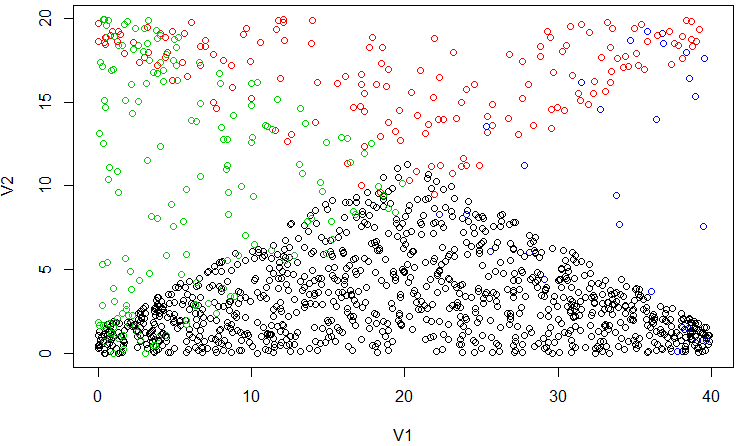}
\par\end{centering}
\caption{Examples of a multiclass dataset with fixed imbalance ratio, first
with all negative classes sharing it, then with compounded imbalance.\label{ej_imb}}
\end{figure}

Using those datasets as a basis, 385 additional imbalanced datasets
were created. These new sets present varying imbalance ratios, depending
on the number of classes:
\begin{itemize}
\item For binary datasets, imbalance ratios of 3, 6, 9, 12 and 15 were used, which means one class maintains the same amount of points and the other class has that amount divided by the decided ratio. 
\item For multiclass datasets, 2 types of imbalance situations were considered: one where the imbalanced classes have the same imbalance ratios and another one where those ratios are compounded. In any case, the same IR values (3, 6, 9, 12 and 15) were chosen. The difference between those situations for the same degree of IR can be seen in Figure \ref{ej_imb}.
\end{itemize}

Table \ref{t4} includes a summary of the characteristics of the artificial imbalanced datasets that were used.

\begin{table}[h]
\caption{\label{t4}Outline of the imbalanced artificial datasets.}

\centering{}%

\begin{tabular}{llccr}
\toprule
\multicolumn{2}{c}{{\scriptsize{}Datasets}} & {\scriptsize{}Overl (\%)} & {\footnotesize{}Imbalance ratio} & {\scriptsize{}Cases}\tabularnewline
\toprule

\multirow{3}{*}{{\scriptsize{}2 clas.}} & {\scriptsize{}Circles} & \multirow{2}{*}{} & \multirow{3}{*}{{\footnotesize{}3,6,9,12,15}} & {\scriptsize{}45}\tabularnewline

 & {\scriptsize{}Rectangles} &  &  & {\scriptsize{}45}\tabularnewline

 & {\scriptsize{}Triangles} & {\scriptsize{}0, 5, 10,} &  & {\scriptsize{}45}\tabularnewline
\cline{1-2}  \cline{4-5} 
\multirow{3}{*}{{\scriptsize{}3 clas.}} & {\scriptsize{}Circles pair} & {\scriptsize{}15, 20, 25,} & \multirow{2}{*}{{\scriptsize{}(1,$i,i$), $i$=3,6,9,12,15}} & {\scriptsize{}90}\tabularnewline

 & {\scriptsize{}Circles tog} & {\scriptsize{}30, 50, 100} &  & {\scriptsize{}90}\tabularnewline
 
 & {\scriptsize{}Circles ins} & \multirow{3}{*}{} & {\scriptsize{}(1,$i,i^{2}$), $i$=3,6,9,12,15} & {\scriptsize{}90}\tabularnewline
\cline{1-2} \cline{4-5}
\multirow{2}{*}{{\scriptsize{}4 clas.}} & {\scriptsize{}Rectangles} &  & {\scriptsize{}(1,$i,i,i$), $i$=3,6,9,12,15} & {\scriptsize{}90}\tabularnewline

 & {\scriptsize{}Triangles} &  & {\scriptsize{}(1,$i,i,i^{2}$), $i$=3,6,9,12,15} & {\scriptsize{}90}\tabularnewline
\midrule
\multicolumn{4}{c}{{\scriptsize{}Total}} & {\scriptsize{}585}\tabularnewline
\bottomrule
\end{tabular}

\end{table}

In addition to the collection of artificial datasets and to test the applicability of the results to real situations, several real datasets were selected from the KEEL repository \cite{triguero_keel_2017} and added to the study. Their main features are noted in Table \ref{t5}. They present different numbers of features and classes, and the number of examples does not exceed that of the biggest artificial dataset used. In addition, they have no missing values and all features are numeric.

\begin{table}[!h]
\caption{\label{t5}Characteristics of the chosen real datasets from KEEL.}

\centering{}%
\begin{tabular}{lrrr}
\toprule
{Dataset} & {Examples} & {Features} & {Classes}\tabularnewline
\toprule
{appendicitis} & {106} & {8} & {2}\tabularnewline

{australian} & {690} & {15} & {2}\tabularnewline

{balance} & {625} & {5} & {3}\tabularnewline

{bands} & {365} & {20} & {2}\tabularnewline

{bupa} & {345} & {7} & {2}\tabularnewline

{cleveland} & {297} & {14} & {5}\tabularnewline

{contraceptive} & {1,473} & {10} & {3}\tabularnewline

{ecoli} & {336} & {8} & {8}\tabularnewline

{glass} & {214} & {10} & {6}\tabularnewline

{haberman} & {306} & {4} & {2}\tabularnewline

{hayes-roth} & {160} & {5} & {3}\tabularnewline

{heart} & {270} & {14} & {2}\tabularnewline

{hepatitis} & {80} & {20} & {2}\tabularnewline

{ionosphere} & {351} & {34} & {2}\tabularnewline

{iris} & {150} & {5} & {3}\tabularnewline

{newthyroid} & {215} & {6} & {3}\tabularnewline

{pima} & {768} & {9} & {2}\tabularnewline

{segment} & {2,310} & {20} & {7}\tabularnewline

{vehicle} & {846} & {19} & {4}\tabularnewline

{wine} & {178} & {14} & {3}\tabularnewline

{winequality-red} & {1,599} & {12} & {6}\tabularnewline

{wisconsin} & {683} & {10} & {2}\tabularnewline

{yeast} & {1,484} & {9} & {10}\tabularnewline
\bottomrule
\end{tabular}
\end{table}

\subsection{Classification methodology}\label{subsec:methodology}

The classification task was conducted using a 5-fold cross-validation scheme, both for the artificial and real datasets. To check how different types of classifiers interact with the proposed datasets, 3 different classifiers were chosen for the study.
\begin{itemize}
\item As a representative of the instance-based classifiers, the KNN method from the \texttt{kknn} R package \cite{schliep_kknn_nodate} was chosen. Two different low values of k were used: k=1 and k=3. Choosing higher values of k would favour the negative class/es, which should be avoided. For this algorithm, the data is normalised, no weights are attached and the Euclidean distance is used.
\item As a representative of decision trees, C4.5 was selected from the \texttt{RWeka} package \cite{hornik_weka_classifier_trees_nodate}.
\item As a representative of the probabilistic classifiers, Na\"ive Bayes (from the \texttt{e1071} R package \cite{meyer_naivebayes_nodate}) was chosen, with no Laplace smoothing. 
\end{itemize}

The classification results were evaluated using two performance metrics: the AUC and the geometric mean (GEOM). Both of them are suitable for evaluating the classification performance, even over imbalanced datasets. 

Specifically, the AUC was evaluated using the generalisation explained in \cite{hand_simple_2001}, which can work with multiclass problems and is estimated using Equation \ref{eq:AUC}, where $c$ is the number of classes, and $\hat{A}(i,j)$ is the mean of the probabilities that either a random element of class $i$ has a better chance of being from class $j$ than a random element of class $j$ and the opposite.

\begin{equation}
AUC=\frac{2}{c(c-1)}\underset{i<j}{\sum}\hat{A}(i,j),\label{eq:AUC}
\end{equation}

An adaptation of the geometric mean was also implemented for its use with multiclass problems, using the ratios of right guesses for all the classes. Given A the confusion matrix of the classification results, with elements $a_{i,j}$, the Equation \ref{eq:geom} is the multiclass geometric mean that was used.

\begin{equation}
GEOM(A)=\sqrt[n]{\stackrel[i=1]{n}{\prod}\frac{a_{i,i}}{\stackrel[j=1]{n}{\sum}a_{i,j}}}.\label{eq:geom}
\end{equation}

\section{Results and Discussion on the Morphology-based Complexity Metrics}\label{sec:results}


In this section, the results of the study are presented. Firstly, the effects of the known overlap and class imbalance of the artificial datasets on the classification results are evaluated in Section \ref{subsec:overimb}. Then, the correlations between the classification results and the set of complexity metrics are presented in Section \ref{subsec:relationship}. Finally, the discussion of the results and some lines for future work are shown in Section \ref{subsec:discussion}.

\subsection{Effects of overlap and imbalance on the classification performance}\label{subsec:overimb}

First of all, it must be noted that this section only shows some of the obtained relevant examples (due to space and readability constraints); the full results are available online~\footnote{\url{https://github.com/jdpastri/morphology-metrics}}.

\begin{figure}[H]
    \centering
    \subfloat[]{\label{clas1a}
    \includegraphics[width=.45\linewidth]{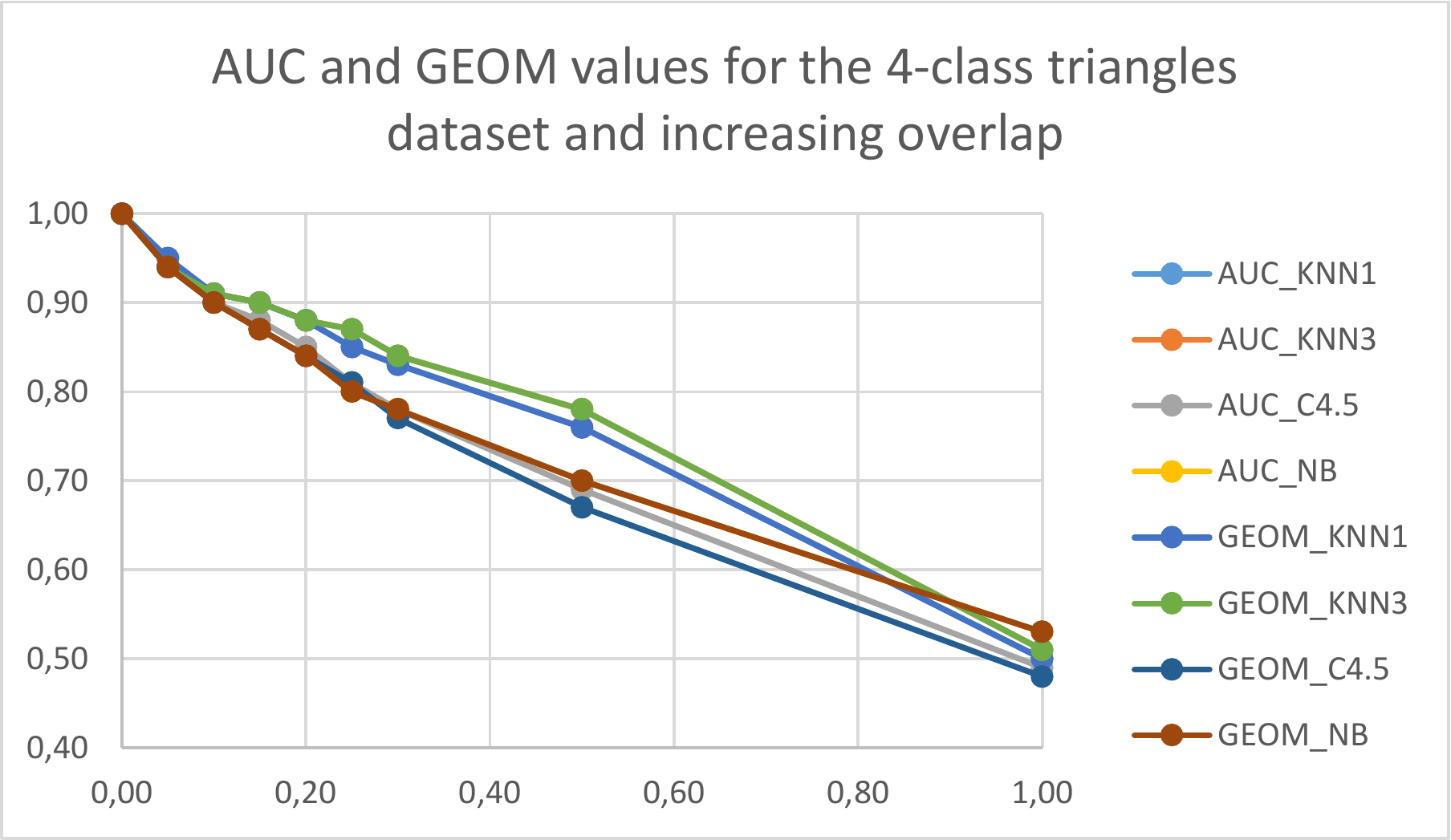}}

    \subfloat[]{\label{clas1b}
    \includegraphics[width=.45\linewidth]{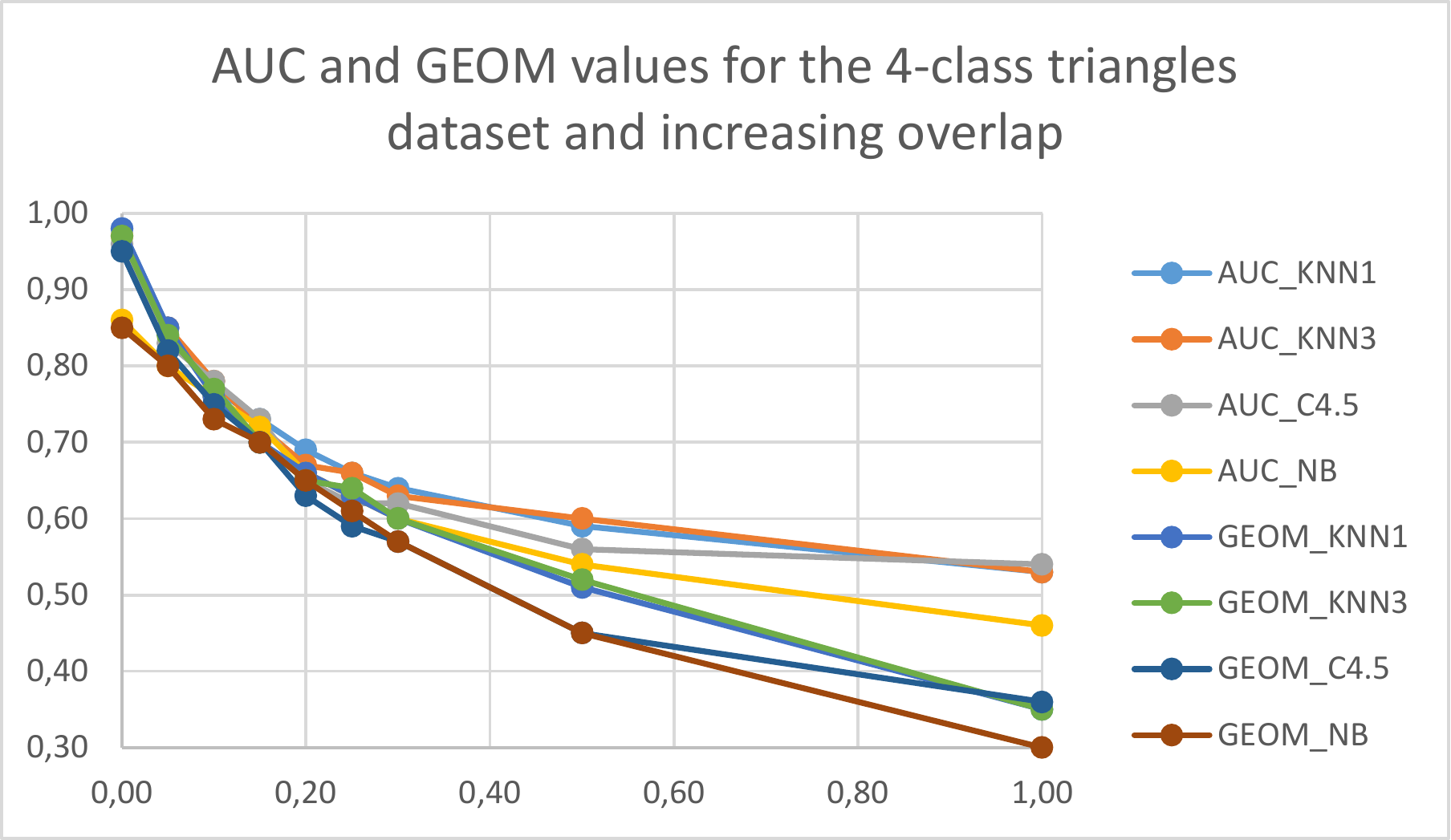}}
\caption{\label{clas1}Examples of the decrease of classification efficacy of the binary circles (a) and 4-class triangles (b) datasets with the increase in overlap.}

\end{figure}

From the classification results obtained when using the artificial datasets, whose overlap ratios are controlled by construction, the fact that increasing class overlap decreases the performance of the classifiers can be observed. Figure \ref{clas1} plots the AUC and GEOM of the classifiers on 2 types of artificial datasets over increasing degrees of overlap. Similar situations happen for the other datasets, although their trends may vary.

\begin{figure}[H]
\centering
    \subfloat[]{\label{clas2a}
    \includegraphics[width=.44\linewidth]{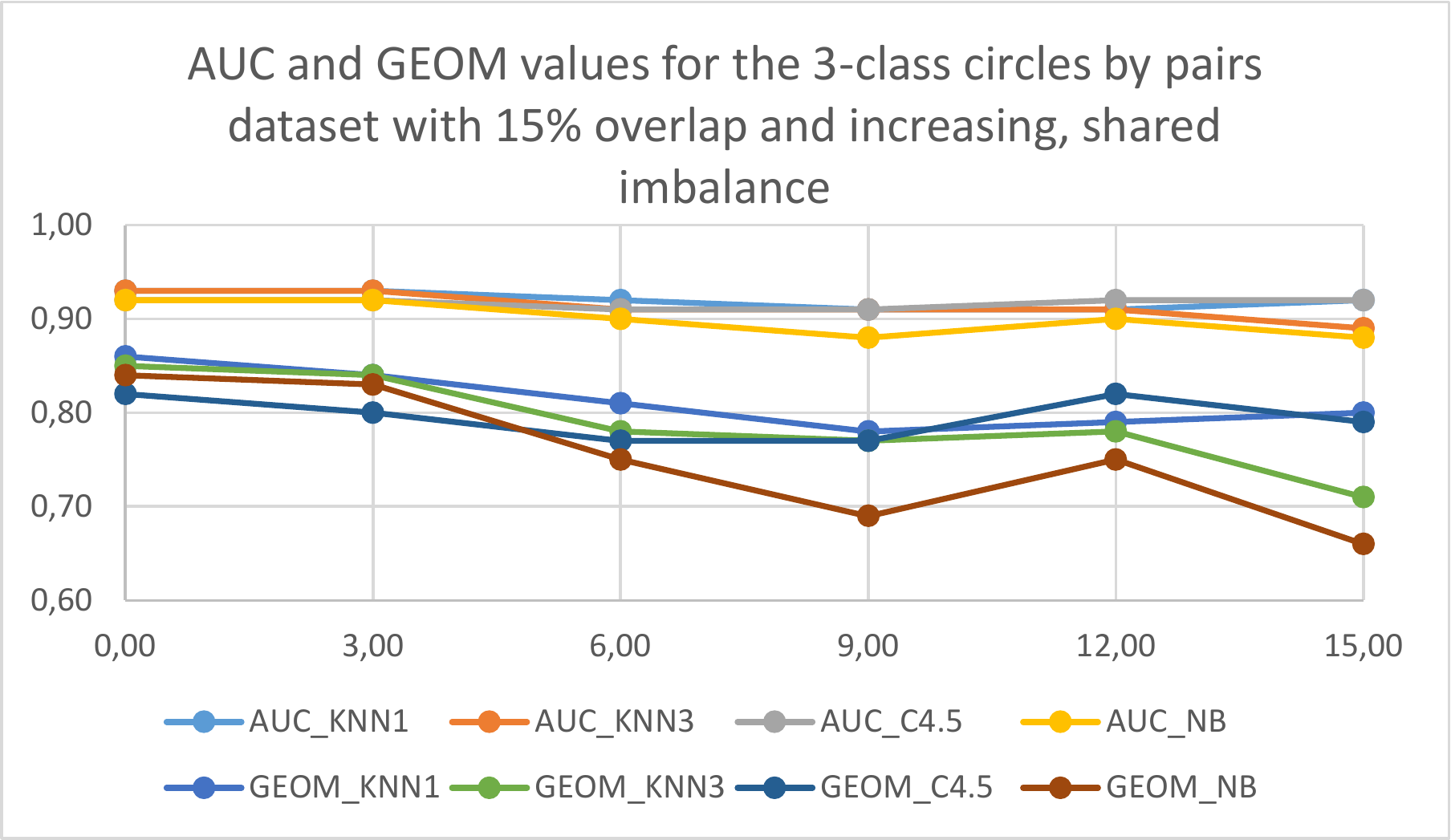}}

    \subfloat[]{\label{clas2b}
    \includegraphics[width=.44\linewidth]{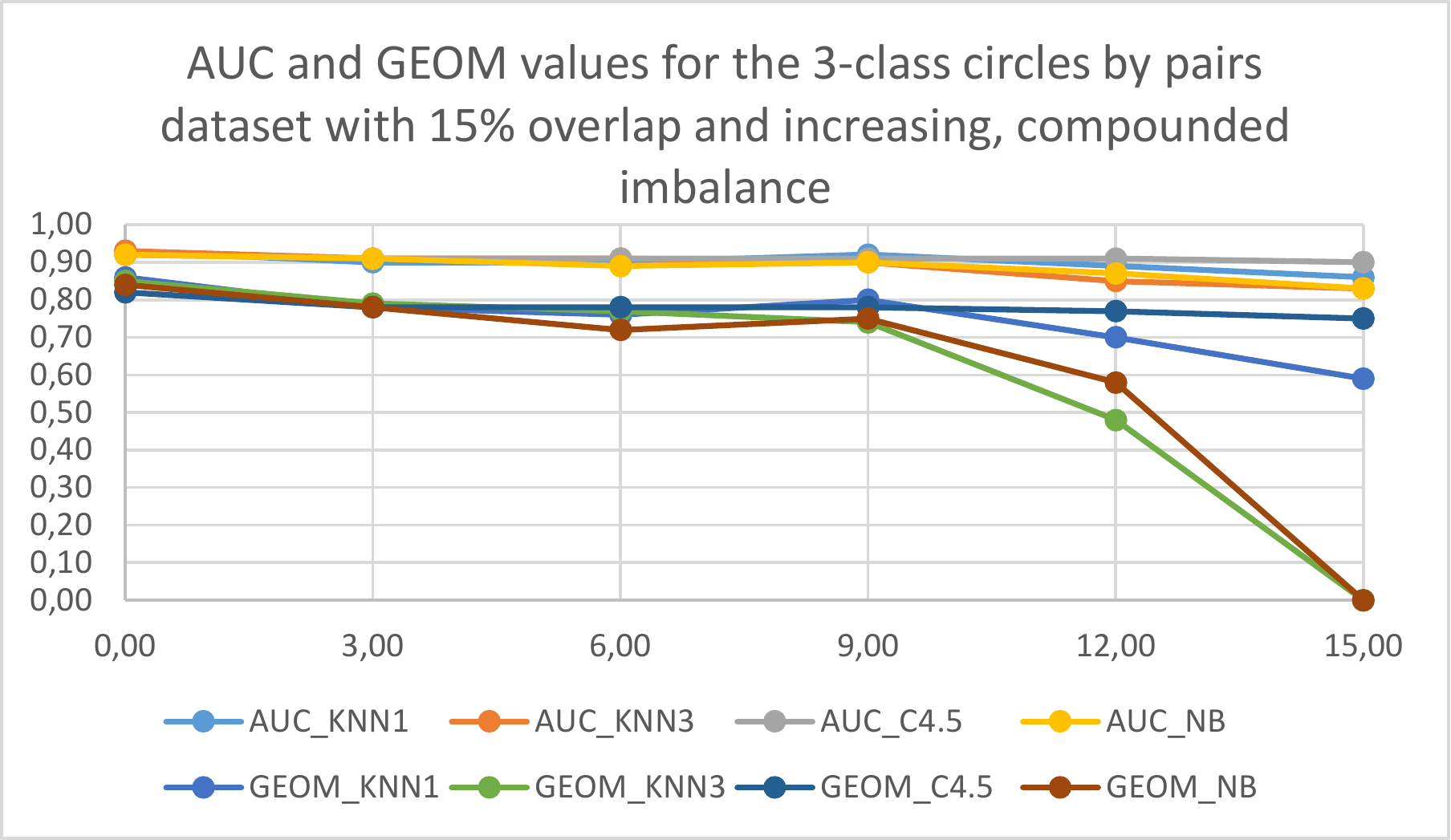}}
\caption{\label{clas2}Examples of the decrease of classification efficacy with the increase in (a) shared and (b) compounded imbalance.}
\end{figure}

Something similar can be observed when only the imbalance ratio is increased, with fixed overlap. Those examples can be seen in Figure \ref{clas2} for the 3-class circles by pairs datasets and different imbalance paradigms. 

\begin{figure}[H]
\begin{centering}
\includegraphics[width=.5\linewidth]{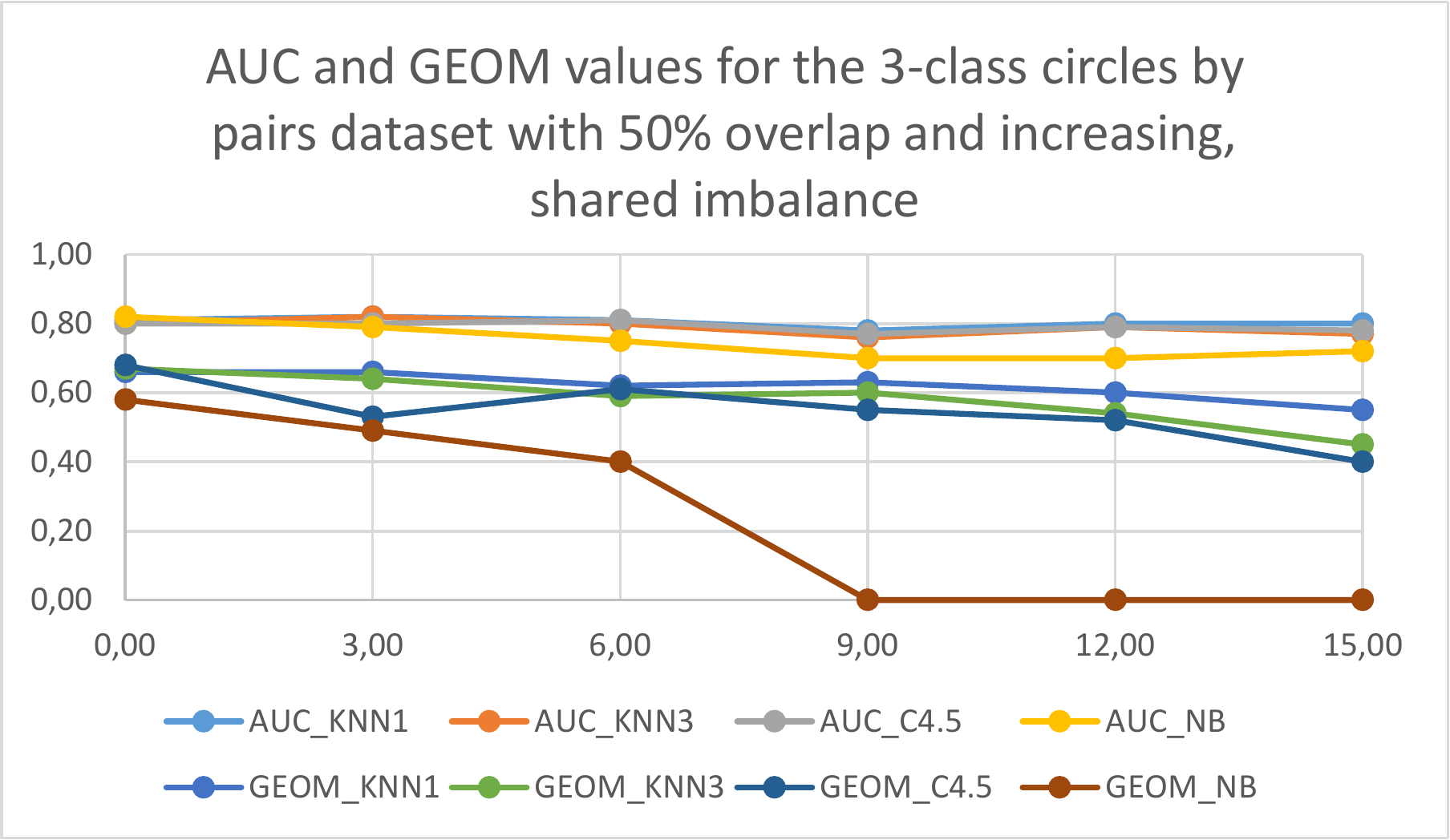}
\par\end{centering}
\caption{\label{clas3}Further decrease of classification efficacy with the increase in overlap and imbalance.}
\end{figure}

Furthermore, this effect is accentuated in the presence of imbalance; that is, an increasing imbalance ratio increases the steepness of the performance fall (see Figure \ref{clas3}, which uses the same type of dataset of Figure \ref{clas2} for contrast). This confirms the suspicion that, even though imbalance and overlap can be detrimental to the performance on their own, the effect is especially notable when both are present simultaneously.

\subsection{Relationship between the classification results and the complexity metrics}\label{subsec:relationship}

One of the main objectives of this study was to evaluate the relationship between the complexity metrics and the classification results. For that purpose, the 4 chosen classifiers (1NN, 3NN, C45 and NB) were used and their performances were compared to the metrics using the Pearson correlation index. Table \ref{t19} includes the mean classification performance per classifier, performance metric and
type of dataset. It also includes the correlations between the classification results, separately for each classifier, performance metric and type of dataset, and the complexity metrics that provided the best overall results. Moreover, the correlation between the different complexity metrics and the theoretical overlap of the artificial datasets is also noted, separately for the balanced and imbalanced datasets.

\begin{table}[!ht]
\caption{\label{t19}Overview of the results obtained with the overall winning
metrics.}

\centering{}%

\begin{tabular}{cclrrrr}
\toprule
\multirow{2}{*}{{Data}} & \multirow{2}{*}{{Perf.}} & \multirow{2}{*}{{Clas}} & {Mean} & \multicolumn{3}{c}{{Correlation with performance}}\tabularnewline
\cline{5-7}
 &  &  & {perf.} & {N3} & {N1} & {$ONB_{avg}^{man}$}\tabularnewline
\toprule

\multirow{9}{*}{\begin{turn}{90}
{Artificial balanced}
\end{turn}} & \multirow{4}{*}{\begin{turn}{90}
{AUC}
\end{turn}} & {1nn} & {.78722} & {-.84287} & {-.83736} & \textbf{-.97159}\tabularnewline

 &  & {3nn} & {.79055} & {-.84654} & {-.84238} & \textbf{-.96945}\tabularnewline
 
 &  & {C4.5} & {.74527} & {-.68445} & {-.70968} & \textbf{-.71784}\tabularnewline

 &  & {NB} & {.71472} & {-.61669} & {-.65413} & \textbf{-.65449}\tabularnewline
\cline{2-7}
 & \multirow{4}{*}{\begin{turn}{90}
{GEOM}
\end{turn}} & {1nn} & {.74916} & {-.89808} & {-.89566} & \textbf{-.99765}\tabularnewline

 &  & {3nn} & {.75305} & {-.90208} & {-.89996} & \textbf{-.99646}\tabularnewline

 &  & {C4.5} & {.69694} & {-.78521} & \textbf{-.8119} & {-.81155}\tabularnewline

 &  & {NB} & {.64861} & {-.61963} & \textbf{-.6641} & {-.6447}\tabularnewline
\cline{2-7}
 & \multicolumn{3}{c}{{Correl. theoretical overlap}} & {.83614} & \textbf{.85709} & {.78833}\tabularnewline
\midrule
\multirow{9}{*}{\begin{turn}{90}
{Artificial imbalanced}
\end{turn}} & \multirow{4}{*}{\begin{turn}{90}
{AUC}
\end{turn}} & {1nn} & {.77047} & {-.76821} & {-.77738} & \textbf{-.91146}\tabularnewline

 &  & {3nn} & {.75842} & {-.74782} & {-.76496} & \textbf{-.88713}\tabularnewline
 
 &  & {C4.5} & {.74817} & {-.64555} & {-.66201} & \textbf{-.74074}\tabularnewline

 &  & {NB} & {.684837} & {-.53964} & {-.60355} & \textbf{-.61906}\tabularnewline
\cline{2-7} 
 & \multirow{4}{*}{\begin{turn}{90}
{GEOM}
\end{turn}} & {1nn} & {.62360} & {-.64221} & {-.63954} & \textbf{-.84805}\tabularnewline

 &  & {3nn} & {.56013} & {-.57571} & {-.57708} & \textbf{-.79894}\tabularnewline

 &  & {C4.5} & {.48070} & {-.55215} & {-.56526} & \textbf{-.74016}\tabularnewline

 &  & {NB} & {.36035} & {-.49918} & {-.56023} & \textbf{-.64383}\tabularnewline
\cline{2-7}
 & \multicolumn{3}{c}{{Correl. theoretical overlap}} & \textbf{.83223}{ } & {.81691 } & {.75632 }\tabularnewline
\midrule
\multirow{8}{*}{\begin{turn}{90}
{Real}
\end{turn}} & \multirow{4}{*}{\begin{turn}{90}
{AUC}
\end{turn}} & {1nn} & {.76782} & \textbf{-.76607} & {-.70171} & {-.59698}\tabularnewline

 &  & {3nn} & {.77304} & \textbf{-.7423} & {-.71645} & {-.58738}\tabularnewline

 &  & {C4.5} & {.77869} & \textbf{-.77106} & {-.6413} & {-.72452}\tabularnewline

 &  & {NB} & {.76782} & \textbf{-.7602} & {-.64488} & {-.61445}\tabularnewline
\cline{2-7} 
 & \multirow{4}{*}{\begin{turn}{90}
{GEOM}
\end{turn}} & {1nn} & {.62304} & {-.6344} & {-.55854} & \textbf{-.69545}\tabularnewline

 &  & {3nn} & {.58304} & {-.65588} & {-.60361} & \textbf{-.71989}\tabularnewline

 &  & {C4.5} & {.60347} & {-.69968} & {-.60392} & \textbf{-.79109}\tabularnewline

 &  & {NB} & {.57565} & {-.64376} & {-.56939} & \textbf{-.6749}\tabularnewline
\bottomrule
\end{tabular}
\end{table}

As can be observed, in most cases, the newly proposed metric $ONB_{avg}^{man}$, that is, the ONB metric using the Manhattan distance and differentiating by class, offered the best agreement with the performance of the classifiers. This is especially true for the artificial datasets, which means that, in a controlled environment, the metric can be a good way to estimate the performance of a classifier.

Regarding the overlap estimation, the correlation between the theoretical overlap and the values of the ONB metric is high, even though N1 and N3 slightly outshine it. These neighbourhood state-of-the-art metrics focus mostly on the boundaries among classes, which are the areas where they overlap, so their overlap estimations obtaining great results makes sense.

Focusing on the behaviour of the classifiers with the real datasets, two different behaviours are observed. When using GEOM, similar results to those observed on the artificial datasets are achieved, with notably better correlation than with the other complexity metrics. There is a difference, however, when using AUC. This situation is most likely tied to the fact that the multiclass AUC is an additive result of the binary AUC results for a dataset, whereas the GEOM metric is multiplicative. This means that GEOM harshly penalises classification results where one or more of the classes are often misclassified. The ONB metrics are able to detect these situations, which makes them better suited for the GEOM results than other complexity metrics.

\subsection{Discussion and Perspectives}\label{subsec:discussion}

This section presents the highlights of the research, both in terms of the metrics with the best performances and the possible lines for future work. 



Using the controlled environment provided by the artificial datasets, the fact that overlap and imbalance prove detrimental to the work of classifiers, specially when simultaneously present, was shown. 

Regarding the relationship between the classification performance metrics and complexity metrics, the ones that best performed belonged to the groups that studied the morphology (ONB) and the neighbourhoods of data points (N3, N1). From them, the $ONB_{avg}^{man}$ metric stands out as the best option with the artificial datasets, especially when using KNN as the classifier (since both are based on distances between points), as well as when using the GEOM classification metric over the real datasets (since
both punish hard-to-classify classes). This allows to emphasize that studying the morphology of datasets is a good way to estimate the complexity of classification problems, since it involves the study of multiple characteristics of the data simultaneously. Furthermore, it also shows that morphology has a high degree of relationship with the theoretical overlap of a dataset, which leads to better decision-making when choosing whether to preprocess datasets or change the classification algorithms for a given classification problem.

The main downside is the speed of obtaining the ONB measures, which involves the creation of a distance matrix from the data points, a computationally expensive process with a quadratic growth given an increase in data points. This could probably be improved further, using heuristics, in order to make the algorithm faster. 

\section{How to adapt ONB to Singular Problems: Prospects}\label{sec:adap}

Even though this study has been focused on standard supervised classification problems (either binary or multiclass and both balanced or imbalanced), there might be benefits to using the proposed morphology metrics on a wider variety of situations, closer to the more complex problems and situations that would appear. Thus, some ideas will now be presented regarding the adaptation of the algorithm (or the problem) so that the ONB metric can be used in singular problems, a particular group of classification problems with more complex inputs or outputs  \cite{prati_2018}.

Specifically, the three main types of singular problems and their adaptation options will be highlighted: multi-label problems in Section \ref{multilabel}, multi-instance problems in Section \ref{multiinstance} and multi-view problems in Section \ref{multiview}. The adaptation towards other singular problems follows similar guidelines to the former case scenarios in Section \ref{others}.

\subsection{Multi-label problems}\label{multilabel}

In multi-label classification problems \cite{herrera_multilabel_2016}, an instance can have multiple labels simultaneously, whereas in traditional classification problems each sample belongs to solely one class. This is the case, for example, of document \cite{katakis_multilabel_2008} or media \cite{boutell_learning_2004} tagging, where ``romantic'', ``science fiction'' or ``comedy'' can be assigned to the same instance. A formal representation of a multi-label problem emerges from the consideration of the set of all possible labels, $\mathcal L=\{l_1,\dots l_p\}$. The prediction space of a multi-label classifier is then $Y= 2^{\mathcal L}\cong\{0,1\}^p$.

In order to perform the multi-label classification, there are 2 main strategies: either transforming the multi-output problem into a series of independent single-output problems (binary, label ranking or ensemble learning problems) whose classification can be achieved using single-output classification algorithms, or adapting a classification algorithm so that
it can work with multi-label data.

As with other classification problems, imbalance can be present among the labels, causing the existence of both minority and majority labels. However, this case is even more complex, as both minority and majority labels can be simultaneously present in a single instance \cite{charte_addressing_2015}.
 
In order to adapt ONB to these problems, different approaches can be used for the inclusions or exclusions inside a ball, centred on a certain point:
\begin{itemize}
\item The new point needs to include, at least, all the labels of the centre.
\item The new point needs to include, at least, one of the labels of the centre.
\item The new point needs to include, at most, some of the labels of the centre (no outside labels allowed).
\item The new point needs to include only non-conflicting labels (for example, points with majority labels cannot be included if the centre only has minority labels). 
\end{itemize}

\subsection{Multi-instance problems}\label{multiinstance}

In this paradigm, instances are given in small groups or ``bags'', both for the training and test sets \cite{herrera_multiple_2016}. Each instance belongs to the same feature space, but each bag can have a different number of them. In more formal notation, if $X$ is the input feature space, a multi-instance classification algorithm would find a map $f:\{s\subset X:s\mbox{ finite}\}\rightarrow Y$. This is the case, for example, of some image classification problems, where said images can be partitioned and classified as a whole \cite{astorino_multiple_2019}.

The aim of the multi-instance classification is to be able to either learn a model that can identify the class of every individual test data point or to identify the global
class of each test bag.

For the classification of the bags, an assumption involving the relationship between the individual instance classes and its bags has to be made. The standard assumption indicates that a bag is positive as long as it contains at least one positive element (it does not matter which, so no individual classification is necessary), and is negative otherwise. However, other assumptions have different conditions or thresholds for the individual instances contained in a bag so that such bag can be classified accordingly. This means, naturally, that if the data point identification is achieved, the bag identification is attained as well (but not conversely). Some methods provide better results for instance-label problems (those centred on instance-level), whereas bag-level or embedding approaches are better suited for bag labelling \cite{carbonneau_2016}.

Regarding this kind of problems, using the bags as points makes the adaptation of the algorithm straightforward. Different approaches
can be taken depending on the paradigm that is used for the bag classification and the selection of the points for the representative of a bag:
\begin{itemize}
\item If any positive point induces a positive ball, the centre of a positive bag can be considered to be one of its positive points (or a combination of them, such as their mean point). Likewise, the centre of a negative bag can be one of the negative points or their mean point. The distance between bags can be measured either between their centres or the closest elements between the bags.
\item Otherwise, the feature vectors can be used separately for the metric, point by point, ignoring the bags.
\end{itemize}

\subsection{Multi-view problems}\label{multiview}

In multi-view problems each instance has a fixed number of feature vectors that can vary in type and format \cite{zhao_multi-view_2017,multiview19}. More precisely, we consider a list of $v$ feature spaces $X_1,\dots X_v$ and a multi-view classifier can thus be represented by a map $f:X_1\times \dots \times X_v\rightarrow Y$. This happens, for example, in disease diagnostics, where the results of different tests must be aggregated to identify the illness.  In these problems, various options can also be taken. The goal is to use the metric in each feature space independently and combine the results afterwards. This combination can follow two different strategies:
\begin{itemize}
\item Two points need to be compatible in all features simultaneously to belong to the same ball.
\item Two points need to be compatible in, at least, one feature to belong to the same ball. 
\end{itemize}

\subsection{Other singular problems}\label{others}

Several existing learning problems are not included in previous typologies, as can be seen in \cite{charte_snapshot_2019}. From them, the most relevant and their associated strategies for ONB adaptation are the following:
\begin{itemize}
\item Multi-dimensional classification, which involves classifying in various fronts simultaneously \cite{Bielza2011705}. The same approach for ONB as with multi-view problems can be taken.
\item Label distribution learning, where labels are assigned in different degrees for each instance \cite{Geng20161734}. The approach for ONB would be similar to that of multi-label problems.
\item Label ranking problems, where a ranking of the labels is given for each instance \cite{Hullermeier20081897}. The approach for ONB would also be similar to that of multi-label problems, but using thresholds of belonging.
\item Multi-target regression, where there are multiple outputs for the regression \cite{BorchaniVBL15}. If the values of the regression are considered independent and get discretised, ONB can be used normally (separately for the different outputs).
\item Ordinal regression, where the outputs are discrete and ordered \cite{Gu20151403}. The approach is similar to that of multi-target regression.
\end{itemize}

\section{Concluding Remarks}\label{sec:remarks}

This paper started with a review on the state-of-the-art complexity metrics, especially focused on the overlap among classes and data morphology. Then, the usefulness of measuring the complexity and overlap of a dataset, to gauge the classification performance that can be attained as well as to identify the need of data preprocessing, was exposed. As per the results obtained, the proposed morphology metrics, ONB, have proved very promising. On the one hand, $ONB_{avg}^{man}$ does achieve a strong correlation with the classification performance metrics; and, on the other hand, it also provides a good estimation of the overlap of a dataset. 

This study has been useful to illustrate that, while little explored, the study of data morphology can be very beneficial for supervised data classification. In particular, for imbalanced datasets with overlap, the ONB complexity metrics are useful new tools that can outdo most state-of-the-art complexity metrics in both complexity and overlap estimation.

As for future work lines, more real datasets and classifiers can be used and a reduction of the computational complexity (using heuristics) would be advisable. Furthermore, extending the work to singular problems using the suggested methodology, and in particular to multi-label, multi-instance and multi-view classification problems, could provide interesting results. 

\section*{Acknowledgements}
This work has been partially supported by the Spanish Ministry of Economy and Competitiveness under project TIN2017-89517-P, including European Regional Development Funds, and the Andalusian regional project P18-TP-5035. This work is part of the PRII2018-02 Intensification Program from the University of Granada and the FPU National Program (Ref. FPU17/04069).

%
\section*{Conflict of Interest}

The authors declare that they have no conflict of interest.

\bibliographystyle{spmpsci}
\bibliography{xampl,paper_1_ONB_prueba_zotero}

\newcommand{\noopsort}[1]{} \newcommand{\printfirst}[2]{#1}
  \newcommand{\singleletter}[1]{#1} \newcommand{\switchargs}[2]{#2#1}
\begin{thebibliography}{10}
\providecommand{\url}[1]{{#1}}
\providecommand{\urlprefix}{URL }
\expandafter\ifx\csname urlstyle\endcsname\relax
  \providecommand{\doi}[1]{DOI~\discretionary{}{}{}#1}\else
  \providecommand{\doi}{DOI~\discretionary{}{}{}\begingroup
  \urlstyle{rm}\Url}\fi

\bibitem{aggarwal_data_2014}
Aggarwal, C.: Data classification: {Algorithms} and applications.
\newblock Data {Classification}: {Algorithms} and {Applications}. Chapman \&
  Hall/CRC (2014).
\newblock \doi{10.1201/b17320}.
\newblock Pages: 675

\bibitem{Ahmed2019249}
Ahmed, M.: Data summarization: a survey.
\newblock Knowledge and Information Systems \textbf{58}(2), 249--273 (2019).
\newblock \doi{10.1007/s10115-018-1183-0}

\bibitem{alejo_hybrid_2013}
Alejo, R., Valdovinos, R.M., García, V., Pacheco-Sanchez, J.H.: A hybrid
  method to face class overlap and class imbalance on neural networks and
  multi-class scenarios.
\newblock Pattern Recognition Letters \textbf{34}(4), 380--388 (2013).
\newblock \doi{10.1016/j.patrec.2012.09.003}

\bibitem{alpaydin_machine_2016}
Alpaydin, E.: Machine {Learning}: {The} {New} {AI}.
\newblock MIT Press (2016)

\bibitem{alshomrani_proposal_2015}
Alshomrani, S., Bawakid, A., Shim, S.O., Fernández, A., Herrera, F.: A
  proposal for evolutionary fuzzy systems using feature weighting: {Dealing}
  with overlapping in imbalanced datasets.
\newblock Knowledge-Based Systems \textbf{73}, 1--17 (2015).
\newblock \doi{10.1016/j.knosys.2014.09.002}

\bibitem{anuradha_self_2014}
{Anuradha}, Gupta, G.: A self explanatory review of decision tree classifiers.
\newblock In: ICRAIE-2014, pp. 1--7 (2014).
\newblock \doi{10.1109/ICRAIE.2014.6909245}

\bibitem{astorino_multiple_2019}
Astorino, A., Fuduli, A., Gaudioso, M., Vocaturo, E.: Multiple instance
  learning algorithm for medical image classification.
\newblock In: SEBD, vol. 2400, pp. 1--8 (2019)

\bibitem{Barboza2017405}
Barboza, F., Kimura, H., Altman, E.: Machine learning models and bankruptcy
  prediction.
\newblock Expert Systems with Applications \textbf{83}, 405--417 (2017).
\newblock \doi{10.1016/j.eswa.2017.04.006}

\bibitem{baumgartner_data_2006}
Baumgartner, R., Somorjai, R.: Data complexity assesment in undersampled
  classification of high dimensional biomedical data.
\newblock Pattern Recognition Letters \textbf{27}, 1383--1389 (2006).
\newblock \doi{10.1016/j.patrec.2006.01.006}

\bibitem{bellinger_2019}
Bellinger, C., Sharma, S., Japkowicz, N., Zaïane, O.: Framework for extreme
  imbalance classification: Swim—sampling with the majority class.
\newblock Knowledge and Information Systems  (2019).
\newblock \doi{10.1007/s10115-019-01380-z}

\bibitem{ben-israel_impact_2020}
Ben-Israel, D., Jacobs, W., Casha, S., Lang, S., Ryu, W.,
  de~Lotbiniere-Bassett, M., Cadotte, D.: The impact of machine learning on
  patient care: {A} systematic review.
\newblock Artificial Intelligence in Medicine \textbf{103} (2020).
\newblock \doi{10.1016/j.artmed.2019.101785}

\bibitem{bergstra_2012}
Bergstra, J., Bengio, Y.: Random search for hyper-parameter optimization.
\newblock Journal of Machine Learning Research \textbf{13}(10), 281--305
  (2012).
\newblock \urlprefix\url{http://jmlr.org/papers/v13/bergstra12a.html}

\bibitem{bernado-mansilla_domain_2005}
Bernadó-Mansilla, E., Ho, T.: Domain of {Competence} of {XCS} {Classifier}
  {System} in {Complexity} {Measurement} {Space}.
\newblock IEEE Transactions on Evolutionary Computation \textbf{9(1)}, 82--104
  (2005).
\newblock \doi{10.1109/TEVC.2004.840153}

\bibitem{Bielza2011705}
Bielza, C., Li, G., LarraÃ±aga, P.: Multi-dimensional classification with
  bayesian networks.
\newblock International Journal of Approximate Reasoning \textbf{52}(6),
  705--727 (2011).
\newblock \doi{10.1016/j.ijar.2011.01.007}

\bibitem{BorchaniVBL15}
Borchani, H., Varando, G., Bielza, C., Larrañaga, P.: A survey on multi-output
  regression.
\newblock Wiley Interdiscip. Rev. Data Min. Knowl. Discov. \textbf{5}(5),
  216--233 (2015)

\bibitem{boutell_learning_2004}
Boutell, M.R., Luo, J., Shen, X., Brown, C.M.: Learning multi-label scene
  classification.
\newblock Pattern Recognition \textbf{37}(9), 1757--1771 (2004).
\newblock \doi{10.1016/j.patcog.2004.03.009}

\bibitem{cano_analysis_2013}
Cano, J.R.: Analysis of data complexity measures for classification.
\newblock Expert Systems with Applications \textbf{40}(12), 4820--4831 (2013).
\newblock \doi{10.1016/j.eswa.2013.02.025}

\bibitem{carbonneau_2016}
Carbonneau, M.A., Cheplygina, V., Granger, E., Gagnon, G.: Multiple instance
  learning: A survey of problem characteristics and applications.
\newblock Pattern Recognition  (2016).
\newblock \doi{10.1016/j.patcog.2017.10.009}

\bibitem{charte_snapshot_2019}
Charte, D., Charte, F., García, S., Herrera, F.: A snapshot on nonstandard
  supervised learning problems: taxonomy, relationships, problem
  transformations and algorithm adaptations.
\newblock Progress in Artificial Intelligence \textbf{8}(1), 1--14 (2019).
\newblock \doi{10.1007/s13748-018-00167-7}

\bibitem{charte_addressing_2015}
Charte, F., Rivera, A.J., del Jesus, M.J., Herrera, F.: Addressing imbalance in
  multilabel classification: {Measures} and random resampling algorithms.
\newblock Neurocomputing \textbf{163}, 3--16 (2015).
\newblock \doi{10.1016/j.neucom.2014.08.091}

\bibitem{cozar_metahierarchical_2019}
Cózar, J., Fernández, A., Herrera, F., Gámez, J.A.: A {Metahierarchical}
  {Rule} {Decision} {System} to {Design} {Robust} {Fuzzy} {Classifiers} {Based}
  on {Data} {Complexity}.
\newblock IEEE Transactions on Fuzzy Systems \textbf{27}(4), 701--715 (2019).
\newblock \doi{10.1109/TFUZZ.2018.2866967}

\bibitem{Das2018HandlingDI}
Das, S., Datta, S., Chaudhuri, B.B.: Handling data irregularities in
  classification: Foundations, trends, and future challenges.
\newblock Pattern Recognit. \textbf{81}, 674--693 (2018)

\bibitem{Fernandez_pareto_2017}
Fernández, A., Carmona, C.J., Del~Jesus, M.J., Herrera, F.: A pareto based
  ensemble with feature and instance selection for learning from multi-class
  imbalanced datasets.
\newblock International Journal of Neural Systems \textbf{27} (2017).
\newblock \doi{10.1142/S0129065717500289}

\bibitem{Fernandez_learning_2018}
Fernández, A., García, S., Galar, M., Prati, R., Krawczyk, B., Herrera, F.:
  Learning from Imbalanced Data Sets.
\newblock Springer (2018).
\newblock \doi{10.1007/978-3-319-98074-4}

\bibitem{feurer_2019}
Feurer, M., Hutter, F.: Hyperparameter Optimization, pp. 3--33.
\newblock Springer (2019).
\newblock \doi{10.1007/978-3-030-05318-5_1}

\bibitem{GALAR2014135}
Galar, M., Fernández, A., Barrenechea, E., Herrera, F.: Empowering difficult
  classes with a similarity-based aggregation in multi-class classification
  problems.
\newblock Information Sciences \textbf{264}, 135–157 (2014).
\newblock \doi{10.1016/j.ins.2013.12.053}

\bibitem{GalarFTSH11}
Galar, M., Fernández, A., Tartas, E.B., Sola, H.B., Herrera, F.: An overview
  of ensemble methods for binary classifiers in multi-class problems:
  Experimental study on one-vs-one and one-vs-all schemes.
\newblock Pattern Recognition \textbf{44}(8), 1761--1776 (2011)

\bibitem{garcia_effect_2015}
Garcia, L.P.F., Carvalho, A.C.P.d.L.F.d., Lorena, A.C.: Effect of label noise
  in the complexity of classification problems.
\newblock Neurocomputing  (2015).
\newblock \doi{10.1016/j.neucom.2014.10.085}

\bibitem{garcia_tutorial_2016}
García, S., Luengo, J., Herrera, F.: Tutorial on practical tips of the most
  influential data preprocessing algorithms in data mining.
\newblock Knowledge-Based Systems \textbf{98}, 1--29 (2016).
\newblock \doi{10.1016/j.knosys.2015.12.006}

\bibitem{Geng20161734}
Geng, X.: Label distribution learning.
\newblock IEEE Transactions on Knowledge and Data Engineering \textbf{28}(7),
  1734--1748 (2016).
\newblock \doi{10.1109/TKDE.2016.2545658}

\bibitem{Gu20151403}
Gu, B., Sheng, V., Tay, K., Romano, W., Li, S.: Incremental support vector
  learning for ordinal regression.
\newblock IEEE Transactions on Neural Networks and Learning Systems
  \textbf{26}(7), 1403--1416 (2015).
\newblock \doi{10.1109/TNNLS.2014.2342533}

\bibitem{gupta2014training}
Gupta, M.R., Bengio, S., Weston, J.: Training highly multiclass classifiers.
\newblock The Journal of Machine Learning Research \textbf{15}(1), 1461--1492
  (2014)

\bibitem{hand_simple_2001}
Hand, D.J., Till, R.J.: A {Simple} {Generalisation} of the {Area} {Under} the
  {ROC} {Curve} for {Multiple} {Class} {Classification} {Problems}.
\newblock Machine Learning \textbf{45}(2), 171--186 (2001).
\newblock \doi{10.1023/A:1010920819831}

\bibitem{herrera_multilabel_2016}
Herrera, F., Charte, F., Rivera, A.J., Jesus, M.J.d.: Multilabel
  {Classification} : {Problem} {Analysis}, {Metrics} and {Techniques}.
\newblock Springer International Publishing (2016).
\newblock \doi{10.1007/978-3-319-41111-8}

\bibitem{herrera_multiple_2016}
Herrera, F., Ventura, S., Bello, R., Cornelis, C., Zafra, A.,
  Sánchez-Tarragó, D., Vluymans, S.: Multiple {Instance} {Learning}:
  {Foundations} and {Algorithms}.
\newblock Springer International Publishing (2016).
\newblock \doi{10.1007/978-3-319-47759-6}

\bibitem{ho_complexity_2002}
Ho, T.K., Basu, M.: Complexity measures of supervised classification problems.
\newblock IEEE Transactions on Pattern Analysis and Machine Intelligence
  \textbf{24}(3), 289--300 (2002).
\newblock \doi{10.1109/34.990132}

\bibitem{hoekstra_nonlinearity_1996}
Hoekstra, A., Duin, R.: On the nonlinearity of pattern classifiers.
\newblock In: Proceedings of 13th {International} {Conference} on {Pattern}
  {Recognition}, vol.~4, pp. 271--275 vol.4 (1996).
\newblock \doi{10.1109/ICPR.1996.547429}.
\newblock ISSN: 1051-4651

\bibitem{hornik_weka_classifier_trees_nodate}
Hornik, K.: Weka\_classifier\_trees function {\textbar} {R} {Documentation}.
\newblock
  \urlprefix\url{https://www.rdocumentation.org/packages/RWeka/versions/0.4-42/topics/Weka_classifier_trees}

\bibitem{Hullermeier20081897}
H\"ullermeier, E., F\"urnkranz, J., Cheng, W., Brinker, K.: Label ranking by
  learning pairwise preferences.
\newblock Artificial Intelligence \textbf{172}(16-17), 1897--1916 (2008).
\newblock \doi{10.1016/j.artint.2008.08.002}

\bibitem{HKV2019}
Hutter, F., Kotthoff, L., Vanschoren, J. (eds.): Automated Machine Learning -
  Methods, Systems, Challenges.
\newblock The Springer Series on Challenges in Machine Learning. Springer
  (2019)

\bibitem{hutter_2019}
Hutter, F., Kotthoff, L., Vanschoren, J.: Automated Machine Learning - Methods,
  Systems, Challenges.
\newblock Springer (2019)

\bibitem{katakis_multilabel_2008}
Katakis, I., Tsoumakas, G., Vlahavas, I.: Multilabel text classification for
  automated tag suggestion.
\newblock Proc. ECML PKDD08 Discovery Challenge p.~9 (2008)

\bibitem{Krawczyk2019601}
Krawczyk, B., Triguero, I., García, S., Woźniak, M., Herrera, F.: Instance
  reduction for one-class classification.
\newblock Knowledge and Information Systems \textbf{59}(3), 601--628 (2019).
\newblock \doi{10.1007/s10115-018-1220-z}

\bibitem{leevy_survey_2018}
Leevy, J., Khoshgoftaar, T., Bauder, R., Seliya, N.: A survey on addressing
  high-class imbalance in big data.
\newblock Journal of Big Data \textbf{5}(1) (2018).
\newblock \doi{10.1186/s40537-018-0151-6}

\bibitem{leyva_set_2015}
Leyva, E., González, A., Pérez, R.: A {Set} of {Complexity} {Measures}
  {Designed} for {Applying} {Meta}-{Learning} to {Instance} {Selection}.
\newblock IEEE Transactions on Knowledge and Data Engineering \textbf{27}(2),
  354--367 (2015).
\newblock \doi{10.1109/TKDE.2014.2327034}

\bibitem{Li_cost_2017}
Li, F., Zhang, X., Zhang, X., Du, C., Xu, Y., Tian, Y.C.: Cost-sensitive and
  hybrid-attribute measure multi-decision tree over imbalanced data sets.
\newblock Information Sciences \textbf{422} (2017).
\newblock \doi{10.1016/j.ins.2017.09.013}

\bibitem{Lorena_analysis_2012}
Lorena, A., Costa, I., Spolaôr, N., de~Souto, M.: Analysis of complexity
  indices for classification problems: Cancer gene expression data.
\newblock Neurocomputing \textbf{75}, 33--42 (2012).
\newblock \doi{10.1016/j.neucom.2011.03.054}

\bibitem{lorena_how_2019}
Lorena, A.C., Garcia, L.P.F., Lehmann, J., Souto, M.C.P., Ho, T.K.: How
  {Complex} is your classification problem? {A} survey on measuring
  classification complexity.
\newblock arXiv:1808.03591  (2019)

\bibitem{luengo_addressing_2011}
Luengo, J., Fernández, A., García, S., Herrera, F.: Addressing data
  complexity for imbalanced data sets: {Analysis} of {SMOTE}-based oversampling
  and evolutionary undersampling.
\newblock Soft Computing \textbf{15}(10), 1909--1936 (2011).
\newblock \doi{10.1007/s00500-010-0625-8}

\bibitem{luengo_2020}
Luengo, J., García-Gil, D., Ramírez-Gallego, S., García, S., Herrera, F.:
  Big Data Preprocessing: Enabling Smart Data.
\newblock Springer (2020).
\newblock \doi{10.1007/978-3-030-39105-8}

\bibitem{luengo_domains_2010}
Luengo, J., Herrera, F.: Domains of competence of fuzzy rule based
  classification systems with data complexity measures: {A} case of study using
  a fuzzy hybrid genetic based machine learning method.
\newblock Fuzzy Sets and Systems \textbf{161}(1), 3--19 (2010).
\newblock \doi{10.1016/j.fss.2009.04.001}

\bibitem{luengo_shared_2012}
Luengo, J., Herrera, F.: Shared domains of competence of approximate learning
  models using measures of separability of classes.
\newblock Information Sciences \textbf{185}(1), 43--65 (2012).
\newblock \doi{10.1016/j.ins.2011.09.022}

\bibitem{luengo_automatic_2015}
Luengo, J., Herrera, F.: An automatic extraction method of the domains of
  competence for learning classifiers using data complexity measures.
\newblock Knowledge and Information Systems \textbf{42}(1), 147--180 (2015).
\newblock \doi{10.1007/s10115-013-0700-4}

\bibitem{luo_2016}
Luo, G.: A review of automatic selection methods for machine learning
  algorithms and hyper-parameter values.
\newblock Network Modeling Analysis in Health Informatics and Bioinformatics
  \textbf{5} (2016).
\newblock \doi{10.1007/s13721-016-0125-6}

\bibitem{Luque2019216}
Luque, A., Carrasco, A., Martín, A., de~las Heras, A.: The impact of class
  imbalance in classification performance metrics based on the binary confusion
  matrix.
\newblock Pattern Recognition \textbf{91}, 216--231 (2019).
\newblock \doi{10.1016/j.patcog.2019.02.023}

\bibitem{Lopez_insight_2013}
López, V., Fernández, A., García, S., Palade, V., Herrera, F.: An insight
  into classification with imbalanced data: Empirical results and current
  trends on using data intrinsic characteristics.
\newblock Information Sciences \textbf{250}, 113–141 (2013).
\newblock \doi{10.1016/j.ins.2013.07.007}

\bibitem{ma_data_2018}
Ma, Y.: Data {Complexity} {Analysis} for {Software} {Defect} {Detection}.
\newblock International Journal of Performability Engineering \textbf{14}
  (2018).
\newblock \doi{10.23940/ijpe.18.08.p5.16951704}

\bibitem{Mahani_2019}
Mahani, A., Baba-Ali, A.: A new rule-based knowledge extraction approach for
  imbalanced datasets.
\newblock Knowledge and Information Systems  (2019).
\newblock \doi{10.1007/s10115-019-01330-9}

\bibitem{manukyan_classification_2016}
Manukyan, A., Ceyhan, E.: Classification of {Imbalanced} {Data} with a
  {Geometric} {Digraph} {Family}.
\newblock J. Mach. Learn. Res.  (2016)

\bibitem{martinez_torres_review_2019}
Martínez~Torres, J., Iglesias~Comesaña, C., García-Nieto, P.J.: Review:
  machine learning techniques applied to cybersecurity.
\newblock International Journal of Machine Learning and Cybernetics
  \textbf{10}(10), 2823--2836 (2019).
\newblock \doi{10.1007/s13042-018-00906-1}

\bibitem{mazurowski_comparative_2011}
Mazurowski, M., Malof, J., Tourassi, G.: Comparative analysis of instance
  selection algorithms for instance-based classifiers in the context of medical
  decision support.
\newblock Physics in Medicine and Biology \textbf{56}(2), 473--489 (2011).
\newblock \doi{10.1088/0031-9155/56/2/012}

\bibitem{meyer_naivebayes_nodate}
Meyer, D.: {naiveBayes} function {\textbar} {R} {Documentation}.
\newblock
  \urlprefix\url{https://www.rdocumentation.org/packages/e1071/versions/1.7-2/topics/naiveBayes}

\bibitem{morais_complex_2013}
Morais, G., Prati, R.C.: Complex {Network} {Measures} for {Data} {Set}
  {Characterization}.
\newblock In: 2013 {Brazilian} {Conference} on {Intelligent} {Systems}, pp.
  12--18 (2013).
\newblock \doi{10.1109/BRACIS.2013.11}

\bibitem{moran_2016}
Morán-Fernández, L., Bolón-Canedo, V., Alonso-Betanzos, A.: Can
  classification performance be predicted by complexity measures? a study using
  microarray data.
\newblock Knowledge and Information Systems  (2016).
\newblock \doi{10.1007/s10115-016-1003-3}

\bibitem{orriols-puig_documentation_2010}
Orriols-Puig, A., Macia, N., Ho, T.K.: Documentation for the data complexity
  library in {C}++.
\newblock Universitat Ramon Llull, La Salle \textbf{196}, 1--40 (2010)

\bibitem{prati_2015}
Prati, R., Batista, G., Silva, D.: Class imbalance revisited: A new
  experimental setup to assess the performance of treatment methods.
\newblock Knowledge and Information Systems \textbf{45}, 245--279 (2014).
\newblock \doi{10.1007/s10115-014-0794-3}

\bibitem{prati_2018}
Prati, R.C., Luengo, J., Herrera, F.: Emerging topics and challenges of
  learning from noisy data in nonstandard classification: a survey beyond
  binary class noise.
\newblock Knowledge and Information Systems \textbf{60}, 63–97 (2019)

\bibitem{rodriguez_bayesian_2015}
Rodriguez, D., Dolado, J., Tuya, J.: Bayesian concepts in software testing:
  {An} initial review.
\newblock In: A-TEST 2015: Proceedings of the 6th International Workshop on
  Automating Test Case Design, Selection and Evaluation, pp. 41--46 (2015).
\newblock \doi{10.1145/2804322.2804329}

\bibitem{schliep_kknn_nodate}
Schliep, K.: kknn function {\textbar} {R} {Documentation}.
\newblock
  \urlprefix\url{https://www.rdocumentation.org/packages/kknn/versions/1.3.1%20/topics/kknn}

\bibitem{scopus_document_nodate}
Scopus: Document {Search}.
\newblock \urlprefix\url{https://www.scopus.com/search/form.uri?display=basic}

\bibitem{serrano_2018}
Serrano, E., Suárez-Figueroa, M., González-Pachón, J., Gomez-Perez, A.:
  Toward proactive social inclusion powered by machine learning.
\newblock Knowledge and Information Systems \textbf{58} (2018).
\newblock \doi{10.1007/s10115-018-1230-x}

\bibitem{Shwartz_2014}
Shalev-Shwartz, S., Ben-David, S.: Understanding Machine Learning: From Theory
  to Algorithms.
\newblock Cambridge University Press, USA (2014)

\bibitem{singh_multiresolution_2003}
Singh, S.: Multiresolution {Estimates} of {Classification} {Complexity}.
\newblock IEEE Trans. Pattern Anal. Mach. Intell.  (2003).
\newblock \doi{10.1109/TPAMI.2003.1251146}

\bibitem{multiview19}
Sun, S., Mao, L., Dong, Z., Wu, L.: Multiview Machine Learning, 1st edn.
\newblock Springer (2019).
\newblock \doi{10.1007/978-981-13-3029-2}

\bibitem{saez_predicting_2013}
Sáez, J.A., Luengo, J., Herrera, F.: Predicting noise filtering efficacy with
  data complexity measures for nearest neighbor classification.
\newblock Pattern Recognition \textbf{46}(1), 355--364 (2013).
\newblock \doi{10.1016/j.patcog.2012.07.009}

\bibitem{tanwani_classification_2010}
Tanwani, A.K., Farooq, M.: Classification {Potential} vs. {Classification}
  {Accuracy}: {A} {Comprehensive} {Study} of {Evolutionary} {Algorithms} with
  {Biomedical} {Datasets}.
\newblock In: J.~Bacardit, W.~Browne, J.~Drugowitsch, E.~Bernadó-Mansilla,
  M.V. Butz (eds.) Learning {Classifier} {Systems}, Lecture {Notes} in
  {Computer} {Science}, pp. 127--144. Springer, Berlin, Heidelberg (2010).
\newblock \doi{10.1007/978-3-642-17508-4_9}

\bibitem{triguero_keel_2017}
Triguero, I., González, S., Moyano, J.M., García, S., Alcalá-Fdez, J.,
  Luengo, J., Fernández, A., Jesús, M.J.d., Sánchez, L., Herrera, F.: {KEEL}
  3.0: {An} {Open} {Source} {Software} for {Multi}-{Stage} {Analysis} in {Data}
  {Mining}.
\newblock International Journal of Computational Intelligence Systems
  \textbf{10}(1), 1238--1249 (2017).
\newblock \doi{10.2991/ijcis.10.1.82}

\bibitem{Vuttipittayamongkol_2020}
Vuttipittayamongkol, P., Elyan, E.: Neighbourhood-based undersampling approach
  for handling imbalanced and overlapped data.
\newblock Information Sciences \textbf{509}, 47--70 (2020).
\newblock \doi{10.1016/j.ins.2019.08.062}

\bibitem{wojciechowski_difficulty_2017}
Wojciechowski, S., Wilk, S.: Difficulty {Factors} and {Preprocessing} in
  {Imbalanced} {Data} {Sets}: {An} {Experimental} {Study} on {Artificial}
  {Data}.
\newblock Foundations of Computing and Decision Sciences \textbf{42}(2),
  149--176 (2017).
\newblock \doi{10.1515/fcds-2017-0007}

\bibitem{zhao_multi-view_2017}
Zhao, J., Xie, X., Xu, X., Sun, S.: Multi-view learning overview: {Recent}
  progress and new challenges.
\newblock Information Fusion \textbf{38}, 43--54 (2017).
\newblock \doi{10.1016/j.inffus.2017.02.007}

\bibitem{zhu_semi-supervised_2005-1}
Zhu, X.: Semi-supervised learning with graphs.
\newblock phd, Carnegie Mellon University, USA (2005).
\newblock AAI3179046 ISBN-10: 0542190591

\end{thebibliography}

\appendix

\section{Description of Other Complexity Metrics}\label{appendix}

\subsection{Feature overlap}\label{subsubsec:overlap}

This group of metrics assesses the capability of features for the discerning of the classes. If there is at least one feature with low overlap of different classes, the classification should be easier and, thus, obtain better results. The same works for combinations of features and areas of the n-dimensional space of each dataset.
Five different metrics can be enumerated:
\begin{itemize}
\item F1: this is the Maximum Fisher’s Discriminant Ratio, which measures the ease to separate the classes using the features (columns) of the data. It compares the dispersion inside each class and the dispersion of the classes. Higher values indicate less overlapping features and, thus, a less complex dataset.
\item F1v: this is the Directional Vector Maximum Fisher’s Discriminant Ratio from \cite{orriols-puig_documentation_2010}. It complements F1, looking for the hyperplane, generated by a vector, that best discerns the classes instead of using the separate features. Greater values indicate a lower complexity.
\item F2: this metric estimates the volume of the overlapping region, using the ratio between the value limits of each class for each feature and the full range of said feature, and multiplying them to get the ratio of volume overlap. Greater values indicate more overlap, which increases the complexity.
\item F3: it is the Maximum Individual Feature Efficiency, which is the biggest ratio of points outside the overlapping region of a feature and the total number of points. Greater values indicate less complexity.
\item F4: this is the Collective Feature Efficiency from \cite{orriols-puig_documentation_2010}. It is based on an iterative use of F3 over the dataset, each time
choosing the most efficient feature and setting aside the non-overlapped points of that feature, until there are no more points or features. F4 indicates the ratio of points that have been discerned over the total number of points, so greater values of F4 signal lower complexity.
\end{itemize}

\subsection{Linearity}\label{subsubsec:linearity}

These measures assess the ease of separability of the different classes by hyperplanes, which would lead to easier classification. There are three main metrics in this category:
\begin{itemize}
\item L1: it is a measure of the Sum of the Error Distance by Linear Programming. After using the linear classifier, the total error distance of the misclassified points to their closest hyperplane is computed and divided by the total number of points. The bigger this ratio, the bigger the L1 will be, indicating more complexity.
\item L2: this is the Error Rate of the Linear Classifier, that is, the number of misclassified points divided by the total number of points.
The bigger the L2, the more complex a problem will be. 
\item L3: it is the Non-Linearity of a Linear Classifier, from \cite{hoekstra_nonlinearity_1996}. For this metric, new points are generated using pairs of points sharing
a class, and they are classified using the initial data as the training set for the generation of the classification model. L3 is the ratio of the misclassified points from those interpolated. A higher value of L3 indicates more complex boundaries and problems.
\end{itemize}

\subsection{Dimensionality}\label{subsubsec:dimensionality}

This set of measures reflect the data sparsity that can appear from a high dimensionality. When a dataset has low-density or even void areas, the model might fail to correctly classify new data there. Three metrics stand out:
\begin{itemize}
\item T2: it is the Average Number of Features per Dimension, that is, the number of features of the dataset divided by the number of points. Greater values indicate less points per feature, which will cause sparsity and a higher complexity.
\item T3: it is the Average number of PCA Dimensions per points. This is the division of the dimensionality of the PCA selected attributes over the number of points of the dataset. Greater values signal a higher complexity.
\item T4: this is the Ratio of the PCA Dimension to the Original Dimension, which is the division of the dimensionality of the PCA selected attributes over the original dimensionality. Greater values indicate that more features are necessary to explain the data variability and, usually, higher complexity. 
\end{itemize}

\subsection{Class balance}\label{subsubsec:balance}

These metrics measure the differences in the number of elements in each class, which could favour the classification of the predominant class. The two most common metrics are:

C1: it is the Entropy of Class Proportions. The higher the value (closer to 1), the most balanced the dataset will be, which usually indicates lower complexity.

C2: this is the Imbalance Ratio, using the multiclass modification in \cite{tanwani_classification_2010}. It has the value ``0'' for balanced problems, and higher values (up to 1) indicate more imbalance.

\subsection{Network properties}\label{subsubsec:network}

These metrics study graph properties of the data, after using the distances between data points to generate it. To this end, each point becomes a node and the instructions in \cite{morais_complex_2013}, \cite{garcia_effect_2015} and \cite{zhu_semi-supervised_2005-1} are followed in order to decide the edges, which only join close points that belong to the same class. The three basic metrics are the following: 

\begin{itemize}
\item Density: it is the Average Density of the Network, which is obtained from the ratio between the number of edges of the graph and the maximum possible amount of edges for that graph. The more edges, the lower the complexity.
\item Clustering Coefficient: this metric is derived from the mean of the ratio of edges between each point and its neighbours and the maximum number of edges between them, for every point. It signals the tendency to create cliques. The higher the value, the most complex the dataset.
\item Hubs: this is the Mean Hub Score of the graph. The hub score measures the importance of each node from both its connections and their hub scores, in an iterative way. The higher the value, the most complex the dataset.
\end{itemize}

\end{document}